%% file: main.tex
\documentclass[opre, nonblindrev]{informs3}
\newtheorem{prop}{Proposition}
\newtheorem{lem}{Lemma}
\newtheorem{thm}{Theorem}

\usepackage{amsmath}
\usepackage{graphicx}
\usepackage{bm}
\usepackage{amssymb}
\usepackage{algpseudocode}
\usepackage{algorithm}
\usepackage{booktabs}
\usepackage{threeparttable}
\usepackage{subcaption}
\usepackage{booktabs}
\usepackage{float} %设置图片浮动位置的宏包
\usepackage{geometry}
\usepackage{enumitem}
\usepackage{hyperref}

\newcommand\mm{^{(t)}}
\newcommand\mmm{^{(t+1)}}
\newcommand\bP{\bm\Phi}
\newcommand\bPs{\bm\Psi}

\newcommand\bX{\bm\Xi}

\newcommand\bC{{\bf{C}}}

\newcommand\bH{{\bf{H}}}

\newcommand\bT{\bm\Theta}
\newcommand\bbT{\bm{T}}

\newcommand\bG{\bm\Gamma}

\newcommand\tr{{\rm tr}}

\newcommand\nmin{N_{\min}}

\renewcommand\t{^{\top}}
\newcommand\la{\lambda_1}
\newcommand\lb{\lambda_2}
\newcommand\X{\mathbf X}
\newcommand\x{\mathbf x}
\newcommand\y{\mathbf y}
\newcommand\Y{\mathbf Y}
\newcommand\e{\bm \varepsilon}
\newcommand\E{\mathbf E}
\newcommand\bE{{\bf\mathcal{E}}}
\newcommand\I{\mathbf{I}}
\newcommand\R{\mathbb{R}}

\DoubleSpacedXII

\usepackage{endnotes}
\let\footnote=\endnote

\usepackage{natbib}
\usepackage{multirow}
\usepackage{booktabs}
\usepackage{pgf,tikz-cd,pgfplots}
\bibpunct[, ]{(}{)}{,}{a}{}{,}%
\def\bibfont{\small}%
\def\bibsep{\smallskipamount}%
\def\bibhang{20pt}%
\TheoremsNumberedThrough     % Preferred (Theorem 1, Lemma 1, Theorem 2)
\ECRepeatTheorems

\EquationsNumberedThrough    % Default: (1), (2), ...
\begin{document}
%%%%%%%%%%%%%%%%

%\setcounter{page}{1}
%\DOI{10.1287/xxxx.0000.0000}%


\RUNTITLE{Row-wise Fusion Regularization: An Interpretable Personalized Federated Learning Framework in Large-Scale Scenarios}

%\TITLE{{\large{Asymptotic Optimality of Projected Inventory Level    Policies  for Perishable Inventory Systems with Positive Lead Times}}}
\TITLE{{\large{Row-wise Fusion Regularization: An Interpretable Personalized Federated Learning Framework in Large-Scale Scenarios}}}

\ARTICLEAUTHORS{%
\AUTHOR{Runlin Zhou}
\AFF{Department of Statistics, University of Science and Technology of China, 
%\EMAIL{zrl1644@mail.ustc.edu.cn},
}
\AUTHOR{Letian Li\footnote{Correspondence to: Letian Li (lilt@mail.ustc.edu.cn)}}
\AFF{Department of Statistics, University of Science and Technology of China,
%\EMAIL{ec4409@stern.nyu.edu}
}
\AUTHOR{Zemin Zheng}
\AFF{Department of Statistics, University of Science and Technology of China,
%\EMAIL{chixiang.chen@som.umaryland.edu}
}
} % end of the block

\ABSTRACT{%
% Enter your abstract
We study personalized federated learning for multivariate responses where client models are heterogeneous yet share variable\mbox{-}level structure. Existing entry\mbox{-}wise penalties ignore cross\mbox{-}response dependence, while matrix\mbox{-}wise fusion over\mbox{-}couples clients. We propose a \emph{Sparse Row-wise Fusion} (SROF) regularizer that clusters row vectors across clients and induces within\mbox{-}row sparsity, and we develop \textbf{RowFed}, a communication\mbox{-}efficient federated algorithm that embeds SROF into a linearized ADMM framework with privacy\mbox{-}preserving partial participation.
Theoretically, we establish an oracle property for SROF—achieving correct variable\mbox{-}level group recovery with asymptotic normality—and prove convergence of RowFed to a stationary solution. Under random client participation, the iterate gap contracts at a rate that improves with participation probability.
Empirically, simulations in heterogeneous regimes show that RowFed consistently lowers estimation and prediction error and strengthens variable\mbox{-}level cluster recovery over NonFed, FedAvg, and a personalized matrix\mbox{-}fusion baseline. A real\mbox{-}data study further corroborates these gains while preserving interpretability. Together, our results position row\mbox{-}wise fusion as an effective and transparent paradigm for large\mbox{-}scale personalized federated multivariate learning, bridging the gap between entry\mbox{-}wise and matrix\mbox{-}wise formulations.
}%

%\KEYWORDS{deterministic inventory theory; infinite linear programming duality;
%  existence of optimal policies; semi-Markov decision process; cyclic schedule}

\KEYWORDS{Personalized federated learning, Row-wise fusion regularization, Multivariate modeling, Sparse estimation} %, large unit penalty cost}
%\HISTORY{ }

\maketitle
%%%%%%%%%%%%%%%%%%%%%%%%%%%%%%%%%%%%%%%%%%%%%%%%%%%%%%%%%%%%%%%%%%%%%%

% Samples of sectioning (and labeling) in OPRE
% NOTE: (1) \section and \subsection do NOT end with a period
%       (2) \subsubsection and lower need end punctuation
%       (3) capitalization is as shown (title style).
%
%\section{Introduction.}\label{intro} %%1.
%\subsection{Duality and the Classical EOQ Problem.}\label{class-EOQ} %% 1.1.
%\subsection{Outline.}\label{outline1} %% 1.2.
%\subsubsection{Cyclic Schedules for the General Deterministic SMDP.}
%  \label{cyclic-schedules} %% 1.2.1
%\section{Problem Description.}\label{problemdescription} %% 2.

% Text of your paper here
%%%%%%%%

%-------------------------------------------------
\section{Introduction}\label{sec:int}
Multivariate learning has become increasingly prevalent in modern applications where models must handle not just a single environment but a sequence of related tasks. In such large-scale multivariate analysis problems, the sheer number of parameters to be estimated can be overwhelming, often necessitating the use of distributed data from multiple clients. Examples include personalized recommendation systems, adaptive pricing engines, and treatment policy optimization in healthcare. In these federated multivariate learning settings, the central challenge lies in efficiently cooperating information and leveraging shared structure across tasks to accelerate learning in each local environment. While a wide range of multivariate methods—such as multiresponse regression, vector autoregression, principal component analysis, and factor analysis—have been well-studied in single-task settings \citep{chen2022fast}, it remains an open question how to extend these benefits to multi-client/multi-task federated frameworks where data heterogeneity and communication constraints are critical considerations.

Recent progress in personalized federated learning (PFL) has begun to address this challenge(\citealp{arivazhagan2019federated}, \citealp{huang2021personalized}). For instance, \citep{liu2025robust} introduced a robust personalized federated learning framework with sparse penalization (PerFL-RSR), which combines Huber loss for heavy-tailed noise robustness with client-wise fusion regularization to capture similarity across models. However, this approach focuses on scalar-wise regression weights and does not generalize naturally to multivariate learning problems. On the other hand, \citep{li2025personalized} proposed Personalized Federated Learning on Large-scale Association Networks, which models matrix-wise relationships between clients via structured low-rank representations. Yet, such matrix-level recovery assumes an unrealistic global prior that lacks interpretability, often forcing artificial clustering of entire coefficient matrices and leading to degraded estimation accuracy in heterogeneous systems. These limitations motivate the need for a more flexible, interpretable, and scalable framework that can simultaneously perform row-wise variable selection and capture cross-client structure in high-dimensional multivariate settings.

To this end, this paper introduces a \underline{\textbf{S}}parse \underline{\textbf{Ro}}w-wise \underline{\textbf{F}}usion Regularization (SROF) framework that shifts the modeling perspective from clustering entire parameter matrices to clustering variable-level row vectors—thereby enhancing interpretability and improving predictive accuracy. Building on this structural assumption, we develop a novel \underline{\textbf{Row}}-wise \underline{\textbf{Fed}}erated algorithm, RowFed, which incorporates SROF into a distributed optimization architecture supporting partial client participation and communication-efficient updates. The proposed penalty integrates the idea of fusion regularization \citep{ma2017concave} with a sparse row-wise structure to simultaneously reveal latent relational pathways and perform supervised variable clustering. We further establish the oracle property of the SROF estimator and provide convergence guarantees for the RowFed algorithm. Together with extensive simulations, our results demonstrate both the theoretical soundness and empirical advantages of leveraging row-wise fusion across clients, yielding improved interpretability and accuracy in large-scale personalized federated learning.

\smallskip
\noindent
\textbf{Challenge.} 
Although several existing methods employ row-wise sparse penalties, such as sparse reduced-rank regression (SRRR) \citep{chen2012sparse} and $\ell_2$-structured variable selection (L2SVS) \citep{simila2007input}, extending them to a federated learning framework that remains computationally efficient and privacy-preserving is nontrivial. We propose a new \textit{row-wise fusion penalty} that generalizes the idea of nonconcave penalized likelihood \citep{fan2011nonconcave} from the entry-wise to the vector-wise setting. However, this extension is mathematically challenging because for any vector $v = (v_1, \ldots, v_q)$, the nonadditivity $p_{\lambda}(\|v\|) \neq \sum_{i=1}^q p_{\lambda}(|v_i|)$ prevents decomposition into simple element-wise forms. The vector-wise setting also introduces complex matrix computations that further complicate both theoretical analysis and optimization. Moreover, once the \textit{SROF estimator} is defined, it is essential to design a suitable federated learning algorithm to compute it under two critical constraints: (i) the algorithm must effectively solve the SROF optimization problem and ensure convergence stability; and (ii) during computation, the data pairs $\{X_m, Y_m\}$ stored on each client must never be transmitted to the central server, thereby ensuring \textit{data privacy} in distributed environments. These difficulties make our framework far from a straightforward extension of existing regularization methods and constitute the central technical challenge addressed in this work.

\smallskip
\noindent
\textbf{Our Contributions.} 
This paper makes the following contributions:
\begin{enumerate}
    \item \textbf{Row-wise fusion framework for PFL.} We propose \textit{RowFed}, a personalized federated learning framework that jointly estimates client-specific parameter matrices via row-wise regularization and fusion learning, enabling interpretable variable-level clustering and improved accuracy under heterogeneity.

    % \item \textbf{Row-wise penalty.} We design a \textit{row-wise sparse penalty} that extends nonconcave penalized likelihood \citep{fan2011nonconcave} from the entry-wise to the vector-wise form, unifying sparsity and cross-client fusion within an interpretable structure.
    
    \item \textbf{Sharp theoretical guarantees.} We establish rigorous theoretical properties of the proposed framework. The \textit{SROF} estimator enjoys the oracle property under standard regularity conditions, and the proposed \textit{RowFed} algorithm is proven to converge to a stationary point with bounded iteration error.
    
    \item \textbf{Practical validation.} Through extensive simulations under heterogeneous and large-scale settings, we demonstrate that \textit{RowFed} consistently outperforms benchmark federated and non-federated methods and PerFL-LSMA in estimation accuracy, and variable cluster identification.
\end{enumerate}

Together, these contributions establish row-wise fusion as a new and interpretable direction for personalized federated learning. Our work bridges the gap between entry-wise and matrix-wise fusion frameworks, filling a key methodological void in multivariate federated analysis.

\subsection{Related Work and Our Distinctions}
\smallskip
\noindent
\textbf{Personalized federated learning.} 
Personalized federated learning (PFL) aims to accommodate data heterogeneity across clients by learning personalized models while maintaining global cooperation. Early approaches focused on model aggregation and parameter decoupling to balance personalization and collaboration, such as those based on client-specific regularization or meta-learning strategies (e.g., model interpolation, clustered FL (\citealp{ghosh2020efficient}, \citealp{sattler2020clustered}), and adaptive aggregation). More recent studies have incorporated sparsity-inducing penalties to enhance interpretability and robustness. For example, \citep{liu2025robust} proposed a robust PFL framework with sparse penalization, which integrates a Huber loss and an entry-wise fusion regularizer to capture pairwise similarity among client models under heavy-tailed noise. This method effectively mitigates the influence of outliers and improves stability in scalar-response settings. However, its entry-wise penalty structure limits the ability to capture dependencies among multiple responses or predictors, making it unsuitable for general multivariate learning problems where cross-variable relationships are essential.

\smallskip
\noindent
\textbf{Large-scale multivariate association recovery.} 
Another relevant line of research concerns large-scale multivariate association recovery, which focuses on estimating structured relationships across multiple response variables and high-dimensional predictors(\citealp{bunea2012joint}, \citealp{reinsel2022reduced}). Existing studies, such as \citep{li2025personalized}, developed matrix-wise fusion strategies to recover shared structures in large association networks. These methods model inter-client similarity through low-rank or matrix-level regularization, enabling information sharing across tasks. Nevertheless, such matrix-wise fusion imposes an overly rigid global prior on the entire parameter matrix, which often leads to interpretability loss and potential bias. In heterogeneous environments, this global coupling can result in mis-clustering and degraded estimation accuracy, as the underlying structure of association typically varies at the variable level rather than at the entire matrix level.

In summary, our work bridges these two lines of research by introducing a row-wise fusion framework for personalized federated multivariate learning, which captures structured dependencies beyond entry-wise penalties and avoids the interpretability loss inherent in matrix-wise fusion.

%-------------------------------------------------
\section{Problem Formulation}\label{sec:PF}

In this section, we formulate the model of \textit{Personalized Federated Learning in Large-Scale Scenarios}. The proposed framework aims to jointly estimate client-specific multivariate models while preserving data privacy across distributed systems. We first present the mathematical formulation of our model in Section~\ref{sec:model}, followed by a background discussion in Section~\ref{sec:background}, where we introduce the general paradigm for solving such problems based on \textit{sparse matrix recovery} and \textit{fusion regularization}. In particular, we review existing approaches in this area, including  \citep{liu2025robust} and \citep{li2025personalized}, highlighting their advantages and inherent limitations when applied to our personalized federated multivariate setting. These comparisons motivate our development of a row-wise fusion regularization framework that unifies interpretability, scalability, and statistical rigor in large-scale personalized federated learning.

\subsection{Model}\label{sec:model}

We consider a federation of $M$ clients. Each client owns a local dataset $\mathcal{D}_m=\{\x_{mi},\y_{mi}\}_{i=1}^{n_m}$ for $m=1,...,M$, where $\y=(y_1,...,y_q)\t \in \R^q$ is a $q$-dimensional response vector, $\x=(x_1,...,x_p)\t\in\R^p$ is a $p$-dimensional covariate vector, and the independent random error $\e=(\varepsilon_1,...,\varepsilon_q)\t\in\R^q$ is a $q$-dimensional  vector. We denote by $n=\sum_{m=1}^M n_m$ the entire sample size. For the $m$-th client, we have
\begin{equation*}
\y_{mi}=\bT^{*\t}_m\x_{mi}+\e_{mi},\,i=1,...,n_m,
\end{equation*}

The independent random error $\e_{mi}$ has $\E(\e_{mi}|\x_{mi})=\bf{0}$, and $\E(\|\e_{mi}
\|^{1+\delta}|\x_{mi}) < \infty$. we denote by $\Y_m=(\frac{1}{\sqrt{n_m}}\y_{m1},...,\frac{1}{\sqrt{n_m}}\y_{mn_m})\t \in \R^{n_{m} \times q}$ the response matrix and by $\X_m=(\frac{1}{\sqrt{n_m}}\x_{m1},...,\frac{1}{\sqrt{n_m}}\x_{mn_m})\t \in \R^{n_{m} \times p}$ the covariate matrix, $\bE_m=(\frac{1}{\sqrt{n_m}}\e_{m1},...,\frac{1}{\sqrt{n_m}}\e_{mn_m})\t$. Considering the data heterogeneity, we assume client-specific linear models: 
\begin{equation}\label{model}
\Y_m=\X_m\bT^*_m+\bE_m,\,m=1,...,M,
\end{equation}
where $\bT_m^*=\left(\bT_{m(1)}^{*}, \ldots, \bT_{m(p)}^{*}\right)^{\top} \in R^{p \times q}$ is true parameter matrix for the $m$-th client and might be different across the clients.

\subsection{Background on Federated Learning in Large-scale Multivariate Scenarios}\label{sec:background}
A common paradigm for personalized federated multivariate learning can be abstractly represented as
\[
\min_{\bT_1, \ldots, \bT_M} \; 
\frac{1}{M}\sum_{m=1}^M \mathcal{L}(Y_m, X_m; \bT_m)
\;+\;
\text{(sparse matrix recovery)}
\;+\;
\text{(fusion regularization)}.
\]
This formulation jointly aims at recovering structured sparsity within each client model while learning cross-client similarity through a fusion mechanism. The \textit{sparse matrix recovery} component may adopt various regularizers, such as sparse reduced-rank regression (SRRR) \citep{chen2012sparse}, sparse orthogonal factor regression (SOFAR) \citep{uematsu2019sofar}, or $\ell_2$-structured variable selection (L2SVS) \citep{simila2007input}, each encouraging low-dimensional and interpretable multivariate representations. The \textit{fusion regularization} component, on the other hand, governs how information is shared across clients, and existing studies can generally be classified into two categories: \textit{entry-wise fusion} and \textit{matrix-wise fusion}.

\smallskip
\noindent
\textbf{Entry-wise fusion.}
\citet{liu2025robust} proposed a robust personalized federated learning framework with sparse penalization, formulated as
\[
\min_{\bT_1, \ldots, \bT_M}
\frac{1}{M}\sum_{m=1}^M 
\mathcal{L}_{\mathrm{Huber}}(Y_m, X_m; \bT_m)
+\sum_{m=1}^{M} \sum_{j=1}^{p} \sum_{i=1}^{q}  p_{\lambda_{1}}\left(\|\bT_{m,ij}\|\right)
+\sum_{m \leq m^{\prime}} \sum_{j=1}^{p} \sum_{i=1}^{q} p_{\lambda_{2}}\left(\|\bT_{m,ij}-\bT_{m^{\prime},ij}\|\right),
\]
which integrates a Huber loss for robustness and an entry-wise fusion term to enforce pairwise similarity among clients. This method performs well under univariate or scalar-response settings and provides robustness to outliers. However, its element-wise penalty structure cannot effectively capture dependencies among multiple response or predictor dimensions, making it less suitable for large-scale multivariate learning.

\smallskip
\noindent
\textbf{Matrix-wise fusion.}
\citet{li2025personalized} proposed a matrix-level fusion approach for large-scale association recovery, expressed as
\begin{align*}
& \min_{\bT_1, \ldots, \bT_M}\bigg\{ 
\frac{1}{M}\sum_{m=1}^{M} \| Y_m - X_m \bT_{m} \|_F^2
+ \sum_{m=1}^{M} \lambda_d \| D_m \|_1
+ \sum_{m=1}^{M} \lambda_a p_a (U_m D_m)
+ \sum_{m=1}^{M} \lambda_b p_b (V_m D_m) \\
& + \sum_{m<m'} \lambda_c p_c (\| \bT_{m} - \bT_{m'}\|_F) 
\bigg\},  
\end{align*}
subject to
\[
\bT_{m} = U_m D_m V_m^\top,\quad U_m^\top U_m = I,\quad V_m^\top V_m = I.
\]
This formulation encourages global similarity through low-rank structure and matrix-wise fusion, enabling information sharing across tasks. Nonetheless, it assumes an overly rigid global prior on the entire parameter matrix, implying that different clients’ parameter matrices are clustered in a similar manner—an assumption that is often too restrictive and impractical in real-world applications—which can lead to interpretability loss and biased estimation under heterogeneous client conditions. When the underlying association structure varies across variables, such global coupling may cause misclustering and reduced estimation accuracy.

\smallskip
\noindent
\textbf{Row-wise fusion: a more interpretable and flexible alternative.}
A more interpretable and flexible way to model cross-client similarity is to perform \textit{row-wise fusion}. This approach assumes that, across different clients, certain variables (rows of the coefficient matrices) may share similar effects even when the overall matrix structures differ. For instance, clients may exhibit consistent effects on specific predictors—such as common genetic factors across patient groups or shared features in recommendation systems—while other predictors remain heterogeneous. An illustrative example of such variable-level similarity is given below:
\[
\bT_1 =
\begin{pmatrix}
1 & 1 & 0\\
0 & 0 & 0\\
3 & 3 & 3\\
\end{pmatrix},\quad
\bT_2 =
\begin{pmatrix}
1 & 1 & 0\\
2 & 2 & 0\\
0 & 0 & 0\\
\end{pmatrix},\quad
\bT_3 =
\begin{pmatrix}
0 & 0 & 0\\
2 & 2 & 0\\
3 & 3 & 3\\
\end{pmatrix} \xrightarrow{\text{matrix-wise}}
\widehat{\bT} =
\begin{pmatrix}
2/3 & 2/3 & 0\\
4/3 & 4/3 & 0\\
2 & 2 & 2\\
\end{pmatrix}.
\]
The structures observed in $\bT_1$, $\bT_2$, and $\bT_3$ indicate that certain rows (corresponding to specific variables) exhibit cross-client consistency. 
For example, the first variable vector can be divided into two clusters: $(1, 1, 0)$ for $\{\bT_1, \bT_2\}$ and $(0, 0, 0)$ for $\bT_3$. 
However, matrix-wise fusion always tends to average these differences and aggregate all matrices into a single representation. 
When the regularization parameter is sufficiently large, it may easily force all matrices to collapse into a common estimate $\widehat{\bT}$, leading to a loss of client-level heterogeneity. 
In contrast, a row-wise fusion design captures shared yet diverse variable-level patterns across clients, providing a more interpretable and flexible modeling strategy.

\smallskip
\noindent
In summary, existing methods based on entry-wise or matrix-wise fusion provide important foundations but face intrinsic limitations in modeling heterogeneous multivariate relationships. In the next section, we propose a \textit{row-wise fusion} framework that bridges these two extremes, enabling flexible variable-level clustering and interpretable personalized federated learning under large-scale settings.

%-------------------------------------------------
\section{Row-wise Variable Fusion Analysis: The RowFed Algorithms}

Building upon the row-wise fusion paradigm introduced in the previous section, we propose the \underline{S}parse \underline{Ro}w-wise \underline{F}usion Learning, i.e \textit{SROF} framework to achieve variable-level clustering and structured sparsity in personalized federated learning. To efficiently solve the corresponding optimization problem under distributed and privacy-preserving constraints, we further develop a communication-efficient federated algorithm, termed \textit{RowFed}. The following subsection introduces the notations and reformulated problem setup underlying this framework.

\smallskip
\noindent
\textbf{Notation.}
To measure the model dissimilarity, we introduce a matrix  $\boldsymbol{\Omega}=\mathbf{E} \otimes \mathbf{I}_{p}$ , where  $\mathbf{E}=\left(\mathbf{e}_{i}-\mathbf{e}_{j}, i<j\right)^{\top} \in \mathbb{R}^{\frac{M(M-1)}{2} \times M}$  with the standard basis vectors  $\left\{\mathbf{e}_{i}\right\} \subset \mathbb{R}^{M}$, $\mathbf{I}_{p}$  is the  $p \times p$  identity matrix and  $\otimes$  is the Kronecker product. For any vector  $\mathbf{x}$ and matrix  $\X$, we use  $\|\mathbf{x}\|=\left(\mathbf{x}^{\top} \mathbf{x}\right)^{1 / 2}$  to denote the  $\ell_{2}$ -norm and  $\|\X\|$  to denote the  F-norm. For any positive definite matrix  $\mathbf{H}$ , we denote  $\|\X\|_{\mathbf{H}}^{2}=\tr(\X\t \mathbf{H} \X)$. For any  $p \times p$  matrix  $\mathbf{M}$ , we use  $\varsigma_{\min}(\mathbf{M})$  to denote the smallest eigenvalue of  $\mathbf{M}$  and  $\varsigma_{\max }(\mathbf{M})$  to denote the largest eigenvalue of  $\mathbf{M}$ . Let  $\nabla f(\bT)$  denote the gradient of a function  $f$  with respect to the model weight  $\bT$ . We say  $a_{n} \lesssim b_{n}$  if  $a_{n} \leq c b_{n}$  for some constant  $c$, $a_{n} \gtrsim b_{n}$  if  $a_{n} \geq c^{\prime} b_{n}$  for some constant  $c^{\prime}$ , and  $a_{n} \asymp b_{n}$  when  $a_{n} \leq c b_{n}$  and  $a_{n} \geq c^{\prime} b_{n}$  both hold.

\subsection{Problem reformulation}

We now focus on the optimization formulation of the proposed Sparse Row-wise Fusion Regularization (SROF) estimator:
\begin{equation}\label{obj}
\begin{aligned}
\widehat{\bT}
=& \underset{\bT_1,\dots,\bT_M \in \mathbb{R}^{p\times q}}{\arg\min}
\Bigg\{
\frac{1}{2M}\sum_{m=1}^{M}\frac{1}{n_m}\sum_{i=1}^{n_m}\|
y_{mi}-\bT_m^{\top}x_{mi}\|^2
+\sum_{j=1}^{p}\sum_{m=1}^{M}p_{\lambda_1}(\|\bT_{m(j)}\|) \\
& +\sum_{j=1}^{p}\sum_{m\le m'}p_{\lambda_2}(\|\bT_{m(j)}-\bT_{m'(j)}\|)
\Bigg\},
\end{aligned}
\end{equation}
where $\bT=(\bT_1^{\top},\dots,\bT_M^{\top})^{\top}\in\mathbb{R}^{Mp\times q}$, $\bT_{(j)}$ denotes the $j$-th row of $\bT$, $p_{\lambda_1}(\cdot)$ and $p_{\lambda_2}(\cdot)$ are penalty functions, and $\lambda_1,\lambda_2\ge0$ are tuning parameters controlling the strengths of row sparsity and inter-client fusion, respectively.  

We refer to $\widehat{\bT}$ as the \textit{SROF estimator}. %The number of estimated clusters is denoted by $\widehat{K}$, and the corresponding cluster memberships by $\widehat{\mathcal{G}}$.
The first penalty term in \eqref{obj} enforces row-level sparsity, while the second encourages similarity across clients, balancing personalization and knowledge sharing. For $p_{\lambda_1}(\cdot)$, we adopt nonconcave penalties such as SCAD \citep{fan2001variable} and MCP \citep{zhang2010nearly} to alleviate bias from over-shrinkage and to improve variable clustering accuracy \citep{ma2017concave}.  

To facilitate algorithmic design, let $\mathbf{A}=[\boldsymbol{\Omega}^{\top},\mathbf{I}_{pM}]^{\top}$ and reformulate \eqref{obj} equivalently as
\begin{equation}\label{obj2}
\begin{aligned}
(\widehat{\bT},\widehat{\bP})
=& \arg\min_{\bT,\bP}
\Big\{
f(\bT)+h_{\lambda}(\bP)
\Big\},
\quad \text{subject to } \mathbf{A}\bT=\bP,
\end{aligned}
\end{equation}
where $\bP_{(j)}$ denotes the $j$-th row of $\bP$, 
$h_{\lambda}(\bP)=\sum_{d}p_{\lambda^*}(\|\bP_{(d)}\|)$,
with $\lambda^*=\lambda_1$ if $d>pM(M-1)/2$ and $\lambda^*=\lambda_2$ otherwise, 
and $f(\bT)=\frac{1}{2M}\sum_{m=1}^{M}\|Y_m-X_m\bT_m\|^2$.  

\smallskip
\noindent
\textbf{Augmented lagrangian.}
To solve the optimization problem \eqref{obj2}, we adopt the alternating direction method of multipliers (ADMM) framework and design a tailored version suitable for personalized federated learning under distributed and privacy-preserving constraints.
We form the augmented Lagrangian function for \eqref{obj2} as
\begin{equation}\label{Lmain}
\mathcal{L}_{\rho}(\bT,\bP,\bG)
= f(\bT)+h_{\lambda}(\bP)
+\langle \bG,\mathbf{A}\bT-\bP\rangle
+\frac{\rho}{2}\|\mathbf{A}\bT-\bP\|^2,
\end{equation}
where $\bG$ denotes the set of Lagrange multipliers and $\rho>0$ is a penalty parameter.  
The alternating direction method of multipliers (ADMM) proceeds as follows:
\begin{itemize}
    \item[1)] $\bT$-step: $\bT^{(t+1)} \leftarrow \arg\min_{\bT}\mathcal{L}_{\rho}(\bT,\bP^{(t)},\bG^{(t)})$;
    \item[2)] $\bP$-step: $\bP^{(t+1)} \leftarrow \arg\min_{\bP}\mathcal{L}_{\rho}(\bT^{(t+1)},\bP,\bG^{(t)})$;
    \item[3)] $\bG$-step: $\bG^{(t+1)} \leftarrow \bG^{(t)}+\rho(\mathbf{A}\bT^{(t+1)}-\bP^{(t+1)})$.
\end{itemize}

The Lagrange multiplier update is straightforward. Since the $\bP$-step aggregates information from all clients while the $\bT$-step can be computed in parallel, the $\bP$-step serves as the central coordination step in the federated system. We now detail the updating procedures.

\smallskip
\noindent
\textbf{Update of $\bT$.}
By expanding \eqref{Lmain}, minimizing with respect to $\bT$ is equivalent to solving
\[
\mathcal{G}(\bT,\bP^{(t)},\bG^{(t)})
= f(\bT)+\frac{\rho}{2}\|\mathbf{A}\bT-\bP^{(t)}+\rho^{-1}\bG^{(t)}\|^2.
\]
To preserve data privacy, the data matrices $(X_m,Y_m)$ on each client remain local. 
We therefore adopt a linearized ADMM approach, and at the same time, to avoid matrix inversion during the optimization process, we define a surrogate function:
\(
\widetilde{\mathcal{G}}(\bT;\bT^{(t)},\bG^{(t)},\bP^{(t)})
= f(\bT^{(t)})
+\langle\nabla f(\bT^{(t)}),\bT-\bT^{(t)}\rangle
+\frac{1}{2\tau}\|\bT-\bT^{(t)}\|_{\mathbf{H}}^2
+\frac{\rho}{2}\|\mathbf{A}\bT-\bP^{(t)}+\rho^{-1}\bG^{(t)}\|^2,
\)
where $\mathbf{H}=r\mathbf{I}-\rho\tau\mathbf{A}^{\top}\mathbf{A}$ is positive definite, and $\tau>0$, $r\ge\rho\tau\varsigma_{\max}(\mathbf{A}^{\top}\mathbf{A})+1$. 
Minimizing this with respect to $\bT$ yields
\begin{equation}\label{Tupdate}
\bT^{(t+1)}=r^{-1}\mathbf{H}\bT^{(t)}-r^{-1}\tau\left[\nabla f(\bT^{(t)})-\rho\mathbf{A}^{\top}\bG^{(t)}+\mathbf{A}^{\top}\bP^{(t)}\right].
\end{equation}

\smallskip
\noindent
\textbf{Update of $\bP$.}
The $\bP$-step solves a concave-penalized least squares problem:
\[
\mathcal{F}(\bT^{(t+1)},\bP,\bG^{(t)})
= h_{\lambda}(\bP)
+\frac{\rho}{2}\|\mathbf{A}\bT^{(t+1)}-\bP+\rho^{-1}\bG^{(t)}\|^2.
\]
Let $\mathrm{S}(\mathbf{z},t)=(1-t/\|\mathbf{z}\|)_+\mathbf{z}$ denote the groupwise soft-thresholding operator.  
Define $\bPs_{(d)}^{(t)}=[\mathbf{A}\bT^{(t+1)}+\rho^{-1}\bG^{(t)}]_d$, where the notation $[\cdot]_d$ indicates the $d$-th row.  
Then, for an $\ell_1$ penalty, the update is given by $\bP_{(d)}^{(t+1)}=\mathrm{S}(\bP_{(d)}^{(t)},\lambda^*/\rho)$.  
For MCP and SCAD penalties, the closed-form updates follow standard thresholding forms (\citealt{zhang2010nearly}; \citealt{fan2001variable}).  
For example, for MCP \citep{zhang2010nearly} with $\gamma>1/\rho$, the solution is
\begin{align}\label{Pup1}
\bP_{(d)}^{(t+1)}=
\begin{cases}
\dfrac{\mathrm{S}\!\left(\bf{\Psi}_{(d)}^{(t)}, \lambda / \rho\right)}{1-1 /(\gamma \rho)}, & \text{if } \|\bf{\Psi}_{(d)}^{(t)}\|_{F} \leq \gamma^{\lambda}, \\[6pt]
\bf{\Psi}_{(d)}^{(t)}, & \text{if } \|\bf{\Psi}_{(d)}^{(t)}\|_{F}>\gamma^{\lambda}.
\end{cases}
\end{align}

It is easy to see that our algorithm described above involves no matrix inversion and primarily consists of one-step updates, laying the groundwork for designing a computationally efficient PFL algorithm.

\subsection{Detial about RowFed algorithm}
For a federated system consists of a large number of clients with variable computational capabilities and network conditions, it is impossible for the server to communicate with all the clients in each communication round. To accommodate system heterogeneity, we design an algorithm that allows low client participation rate in each round.

\smallskip
\noindent
\textbf{Design for federated communication.} 
We design to pass the parameters in the federated system. The updates of $\bT\mmm$ requires local data such that they should be calculated on clients. 
The update \eqref{Tupdate} can be decomposed as  $\bT^{(t+1)}=\widetilde{\bT}^{(t)}-r^{-1} \tau \nabla f\left(\bT^{(t)}\right)$ , where
\begin{align}\label{Tup2}
\widetilde{\bT}^{(t)}=r^{-1} \mathbf{H} \bT^{(t)}-r^{-1} \tau\left[-\rho \mathbf{A}^{\top} \bG\mm+\mathbf{A}^{\top} \bP\mm\right] .
\end{align}

Note that,  $\bG\mm$  and  $\bP\mm$  are used in the weights similarity learning and hence stored in the server. Therefore, after uploading $\bT^{(t)}$ to the server, $\widetilde{\bT}^{(t)}$ can be computed on the server and then downloaded to each client. The gradient  $\nabla f\left(\bT^{(t)}\right)$  requires local data such that it should be calculated on clients. The model weight for the  m  th client is located at the  $1+(m-1) p$  to  $m p$  rows in  $\bT^{(t)}$ , that is,  $\bT_{m}^{(t)}=\left(\bT_{(1+(m-1) p)}^{(t)}, \ldots, \bT_{(m p)}\mm\right)^{\top}$ . Then, the model weights could be updated locally as
\begin{align}\label{Tup3}
\bT_{m}^{(t+1)}=\widetilde{\bT}_{m}^{(t)}-r^{-1} \tau \nabla f_{m}\left(\bT_{m}^{(t)}\right) 
\end{align}
where  $f_{m}\left(\bT_{m}\right)=\frac{1}{2M}\|\Y_m-\X_m\bT_m\|^2$ .

\smallskip
\noindent
\textbf{Partial participation for heterogeneous clients.}
To address system heterogeneity, RowFed adopts partial client participation in each communication round. The participation rate dynamically adapts to system conditions such as the number of clients, computational capabilities, and network bandwidth. When the number of clients is small, full participation can be permitted; conversely, when the federation is large or network latency is high, only a random subset of clients is selected for updating, effectively improving communication efficiency and scalability.

\smallskip
\noindent
\textbf{Proposed algorithm.}
At each iteration, the server holds $\bT\mm$, updates the global quantities $\widetilde{\bT}^{(t)}$, $\bP^{(t+1)}$, and $\bG^{(t+1)}$ based on the aggregated information received from clients. Each selected client $m \in S^{(t)}$ downloads its corresponding $\widetilde{\bT}^{(t)}_m$ from the server, performs the local update according to \eqref{Tup3}, and uploads $\bT^{(t+1)}_m$ back to the server. Clients not selected in round $t$ retain their previous estimates, i.e., $\bT^{(t+1)}_{m'} = \bT^{(t)}_{m'}$ for $m'\notin S^{(t)}$. The initial local estimators $\bT^{(0)}_m$ are computed using local data via the oracle initialization scheme. The entire procedure is summarized in Algorithm~\ref{alg1}.

\begin{algorithm}[!t]
	\begin{algorithmic}[1]
		\Require Tuning parameters $1\le L\le\min\{p,q\}$, $\la,\lb\ge0$, $\rho>0$,  number of iteration $T$.
         \State {\bf Initialization:} Set $\bG^{(0)}\gets{\bf 0}$, compute the  $\bT_m^{(0)}$ by using local data and upload to the server.
		\For{$t=0,1,...,T-1$}
		\State {\bf Server:}
            \State \quad\, Update $\widetilde{\bT}\mm$ by \eqref{Tup2}, $\bP\mmm$ by \eqref{Pup1} and update $\bG\mmm$ following the $\bG$-step. 
            \State\quad\, Select $S\mm\in[M]$ and calculate $\widetilde{\bT}\mm_m$ for each $m\in S\mm$.
            \State {\bf Clients} $m\in S\mm$ {\bf do} in parallel:
            \State \quad\, Download $\widetilde{\bT}\mm_m$.
            \State \quad\, Update $\bT_m\mmm$ by \eqref{Tup3}.
            \State \quad\, Upload $\bT_m\mmm$ to the server.
		\EndFor
            \Ensure Return $\bT_m^{(T)}$ for each $m\in[M]$. 
	\end{algorithmic}
	\caption{RowFed}
	\label{alg1}
\end{algorithm}

\smallskip
\noindent
\textbf{Tuning parameter selection strtegy.} 
In the federated system, the client \(m\) uploaded its local loss \(\frac{1}{n_m} \sum_{i=1}^{n_m} \|\mathbf{y}_{mi} - \widehat{\bT}_m^\top \mathbf{x}_{mi}\|^2\) to the server, and the server calculates the GIC-type criterion \citep{fan2013tuning}. For given \(\lambda_1\) and \(\lambda_2\), the GIC is defined:
\[
    \mathrm{GIC}(\lambda) = \log\{\mathrm{SSE}(\lambda)\} + \left\{\log\log(Nq)\log(pq)/N\right\}\sum_{j=1}^p \widehat{K}_j(\lambda),
\]
where $\lambda = (\lambda_1, \lambda_2)$, \(
    \mathrm{SSE}(\lambda) = \frac{1}{M} \sum_{m=1}^M \frac{1}{n_m} \sum_{i=1}^{n_m} \|\mathbf{y}_{mi} - \widehat{\bT}_m^\top \mathbf{x}_{mi}\|^2,
\)
and $\widehat{K}_j(\lambda)$ is the distinct cluster count for variable $j$. Then we specify a grid value of \(\lambda_1, \lambda_2\) . At last, the parameters \(\lambda_1^*\) and \(\lambda_2^*\) with the smallest GIC value are chosen.

\subsection{RowFed Convergence Analysis}
We analyze the convergence behavior of the proposed \textit{RowFed} algorithm and establish conditions under which the iterative updates are guaranteed to converge to a stationary solution of the SROF optimization problem.

\begin{thm}[Convergence of Algorithm 1]
\label{thm:1}
Let assumptions in Lemma 1 and Assumptions 2 hold. If  $\rho^{(t+1)}=\alpha \rho^{(t)}$  and  $\frac{\min \left\{\lambda_{1}^{2}, \lambda_{2}^{2}\right\}}{2}>\mathcal{L}_{0}+\frac{c_{\lambda} \alpha(\alpha+1)}{2 \rho^{(0)}(\alpha-1)}$ , where  $c_{\lambda}$  is a positive constant decided by  $\lambda_{1}$  and  $\lambda_{2}$ . Assume the selection probabilities for clients are larger than a constant  $r_{p} \in(0,1)$ . Then,  $\left\{\bT^{(t)}, \bP\mm, \bG\mm\right\}_{t=1}^{\infty}$  is bounded and there exists a subsequence  $\left\{\bT^{t_{k}}, \bP^{t_{k}}, \bG^{t_{k}}\right\}_{t_{k}=1}^{\infty}$ , which converges to a stationary point that satisfies the KKT conditions. Further,  $\lim _{t \rightarrow \infty}\left\|\bT^{(t+1)}-\bT^{(t)}\right\|_{2}^{2}+   \left\|\bP\mmm-\bP\mm\right\|_{2}^{2}=0$ , and  $\lim _{T \rightarrow \infty} \sum_{t=0}^{T}\left\|\bT^{(t+1)}-\bT^{(t)}\right\|_{2}^{2} \leq   \frac{1}{r_{p}}\left[\frac{\mathcal{L}^{0}}{c^{(0)}}+\frac{\lambda^{2} \alpha(\alpha+1)}{2 c^{(0)} \rho^{0}(\alpha-1)}\right]$ .
\end{thm}

Theorem \ref{thm:1} indicates that the sequence \(\{\bT^{(t)}\}_{t=1}^{\infty} \) converges to a local minimum and converges at a linear speed of \(O(\frac{1}{r_{p}})\)such that a larger participation rate will fasten the convergence.

%-------------------------------------------------
\section{Statistical Consistency: "SROF" Oracle Property}
In this section, we introduce the latent group notation and oracle estimator, and prove that SROF consistently recovers the true row clusters with probability tending to one; moreover, the nonzero components are asymptotically unbiased and normal under standard conditions.

\subsection{Technical Notation}

\smallskip
\noindent
\textbf{Row-wise cluster structure introduction.}  
When unit-specific coefficients share identical values in some components, such as $\bT^{*}_{m(j)} = \bT^{*}_{m'(j)}$, the $j$th covariate exhibits the same effect across the $m$th and $m'$th units. Modeling each data unit independently therefore neglects such shared structures and may lead to a loss of statistical efficiency and estimation accuracy.  

To formally describe this structure, for each $j \in \{1, \ldots, p\}$, we assume that the set of coefficients $\{\bT_{1(j)}, \ldots, \bT_{M(j)}\}$ can be partitioned into $S_j$ disjoint groups $\mathcal{G}^j_1, \ldots, \mathcal{G}^j_{S_j}$ satisfying $\mathcal{G}^j_s \cap \mathcal{G}^j_{s'} = \emptyset$ for $s \neq s'$ and $\cup_{s=1}^{S_j}\mathcal{G}^j_s = \{1,\ldots,M\}$. For each covariate $j$, the coefficients follow two structural patterns:
\begin{itemize}[leftmargin=2em]
\item[(i)] \textbf{(Heterogeneity)} For all $m \in \mathcal{G}^j_s$, $\bT_{m(j)} = \bX^j_s$, where $\bX^j_s$ denotes the common coefficient vector shared within group $\mathcal{G}^j_s$.  
\item[(ii)] \textbf{(Sparsity)} Under the heterogeneous structure, some $\bX^j_s$ may be exactly zero for certain $s \in \{1,\ldots,S_j\}$.
\end{itemize}
Given these assumptions, we jointly model all $M$ units and treat model~\eqref{model} as a \emph{high-dimensional heterogeneous sparse multivariate regression}.

\smallskip
\noindent
\textbf{Notation and oracle estimator.}  
Let $S = \sum_{j=1}^{p} S_j$ be the total number of distinct coefficient groups, and define $S_0 = n_0 = 0$.  
Let  $\boldsymbol{N}=\left\{N_{s}^{j}, j \in\{1, \ldots, p\}, s \in\right.   \left.\left\{1, \ldots, S_{j}\right\}\right\}=\left\{N_{1}, \ldots, N_{s}\right\}$ , where  $N_{s}^{j}=\sum_{k \in \mathcal{G}_{s}^{j}} n_{k}$ , and  $N_{\min }=\min _{j, s} N_{s}^{j}\left(N_{\max }=\max _{j, s} N_{s}^{j}\right)$  represents the true minimum (maximum) sample size among all the groups.
For each $m \in \{1, \ldots, M\}$, let $\mathbf{W}_m$ be a $p \times S$ binary matrix whose $(j,s')$th entry satisfies $W_{mjs'} = 1$ if $m \in \mathcal{G}^j_s$ with $s' = \sum_{j'=0}^{j-1} S_{j'} + s$, and $W_{mjs'} = 0$ otherwise.  
Denote $\mathbf{W}_{Mp\times S} = (\mathbf{W}_1^\top, \ldots, \mathbf{W}_M^\top)^\top$ and $\mathbb{X}_{n\times S} = \mathbf{X}\mathbf{W}$.  
Define the structured parameter space
\(
\mathcal{M}_{\mathcal{G}} = \big\{\bT \in \mathbb{R}^{Mp\times q} : \bT_{m(j)} = \bT_{m'(j)},\ \forall\, m,m' \in \mathcal{G}^j_s,\ j \in [p],\ s \in [S_j]\big\}.
\)
Let $\bX^j_s$ denote the subgroup-specific coefficients for the $j$th covariate.  
When $\bT \in \mathcal{M}_{\mathcal{G}}$, we can express $\bT = \mathbf{W}\bX$, and by rearranging nonzero components before zeros, write $\bX = (\bX_{\mathcal{K}1}^\top, \bX_{\mathcal{K}2}^\top)^\top$, where $\bX_{\mathcal{K}1} \in \mathbb{R}^{s_k \times q}$ and $\bX_{\mathcal{K}2} \in \mathbb{R}^{(S-s_k)\times q}$ correspond to the nonzero and zero sub-blocks, respectively.  
Hence, the true coefficients can be written as $\bX_0 = (\bX_{0\mathcal{K}1}^\top, \mathbf{0}^\top)^\top$ and $\bT_0 = \mathbf{W}\bX_0$. 

The oracle estimator $\widetilde{\bX}^{*} = (\widetilde{\bX}^{* \top}_{\mathcal{K}1}, \widetilde{\bX}^{* \top}_{\mathcal{K}2})^{\top}$ is defined as the least-squares estimator with prior knowledge of $\mathbf{W}$ and $\bX_{0\mathcal{K}2} = 0$, i.e., $\widetilde{\bX}^{*}_{\mathcal{K}2} = 0$ and
\[
\widetilde{\bX}_{\mathcal{K}1}^{*}
= \underset{\bX_{\mathcal{K}1} \in \mathbb{R}^{s_k \times q}}{\arg\min}\,
(2M)^{-1}\big\|\Y - \mathbb{X}_1 \bX_{\mathcal{K}1}\big\|^{2}.
\]

Let $b_n = \min_{j \in [p],\, S_j>1}\min_{s \neq s'} \|\bX_{0s}^j - \bX_{0s'}^j\|$ denote the minimum inter-group coefficient gap for heterogeneous covariates.

\subsection{Theoretical Results}
\smallskip
\noindent
\textbf{Oracle consistency.}
We next establish that the SROF estimator coincides with its oracle counterpart with high probability, ensuring consistent group recovery.
First of all, we demonstrate the convergence rate of the oracle.  
\begin{prop}\label{oracle}
If Conditions (C1)–(C4) in Appendix~\ref{sec:proSROF} hold and $\nmin\gg s_kq$, we have
$$
\widetilde{\bX}^*_{\mathcal{K} 2}=0, \quad\left\|\widetilde{\bX}^*_{\mathcal{K} 1}-\bX_{0 \mathcal{K} 1}\right\|=O_{p}\left(\sqrt{s_kq / N_{\min}}\right)
$$
with probability tending to one as $N_{\min} \rightarrow \infty$.
\end{prop}
The result of Proposition~\ref{oracle} can be readily extended from Theorem 1 in \cite{ma2017concave}, and the proof is therefore omitted here.

\begin{thm}\label{TXor}
Suppose that conditions (C1)–(C5) in Appendix~\ref{sec:proSROF} hold, and that $\min_{1\le j \le q}\|\bX_{0\mathcal{K}1j}\| > a\gamma$, $b_n > a\lambda$, and $\lambda \gg \sqrt{q/N_{\min}}$.  
Then there exists a local minimizer of the SROF objective such that
\[
\Pr\big(\widehat{\bT}(\lambda_{1}, \lambda_{2}) = \widetilde{\bT}^{*}\big) \to 1,
\]
where $\widetilde{\bT}^{*}$ is the oracle estimator associated with $\widetilde{\bX}^{*}$.
\end{thm}

Theorem~\ref{TXor} implies that if the minimum inter-group difference satisfies $b_n \gg 1/\sqrt{N_{\min}}$, the oracle estimator $\widetilde{\bT}^{*}$ coincides with a local minimizer of the SROF objective with probability tending to one. Consequently, the true variable-wise grouping structure can be consistently identified.

\smallskip
\noindent
\textbf{Asymptotic distribution.}
To establish asymptotic normality, we assume regression error matrix $\bE$,whose elements $\{\bE_{i1}, \ldots, \bE_{iq}\}$ are mutually independent for each $i$.  
Under this assumption, the following result holds.
\begin{thm}\label{Xnornew}
The SROF estimator is asymptotically normal:
\[
\widehat{\sigma}_n(a_n)^{-1} a_n^{\top}(\widehat{\bX}_{\mathcal{K}1} - \bX_{0\mathcal{K}1}) c_n
\xrightarrow{D} \mathcal{N}(0,1),
\]
where \(
\widehat{\sigma}_n(a_n) =
\widehat{\sigma}\Big\{a_n^{\top}(\mathbb{X}_1^{\top}\mathbb{X}_1)^{-1}a_n\Big\}^{1/2}\), \(
\widehat{\sigma}^2 = \frac{1}{n}\sum_{m=1}^{M}\sum_{i=1}^{n_m}
\|\mathbf{y}_{mi} - \bT_m^{\top}\mathbf{x}_{mi}\|^2.
\)
\end{thm}

Theorem~\ref{Xnornew} establishes that, under standard regularity conditions, the proposed SROF estimator is asymptotically unbiased and normally distributed, thereby achieving the \emph{oracle property}.

%-------------------------------------------------
\section{Numerical Evaluation and Real Data Application}
In this section, we present comprehensive simulation studies to compare the performance of our proposed federated learning method (RowFed) with commonly used alternatives. We describe the simulation setup, evaluation indicators, and analysis of results under both heterogeneous and homogeneous scenarios.

\subsection{Synthetic Data Experiments}

\smallskip
\noindent
\textbf{Data Generation.}
We simulate $M$ federated data units, each following the multivariate linear model
\begin{equation*}
\Y_m = \X_m \bT^*_m + \bE_m, \quad m = 1, \ldots, M,
\end{equation*}
where $\Y_m \in \mathbb{R}^{n_m \times q}$ is the response matrix, $\X_m \in \mathbb{R}^{n_m \times p}$ the design matrix, and $\bE_m$ the random error. To ensure comparability, all units share the same sample size $n_1 = \cdots = n_M = n$. The covariates are generated with i.i.d. rows from $\mathcal{N}_p(\mathbf{0}, \bm{\Sigma}_x)$, where $\bm{\Sigma}_x = (0.5^{|i-j|})$, and the errors have i.i.d. rows from $\mathcal{N}_q(\mathbf{0}, \bm{\Sigma}_e)$ with $\bm{\Sigma}_e = (0.5^{|i-j|})$. 

To emulate heterogeneous effects, the true coefficient matrices $\bT^*_m$ are constructed by dividing their row vectors into two latent clusters. Specifically, for each row $j$, 
\[
\mathbb{P}(\bT^*_{m(j)} = \mathbf{v}_j^*) = \mathbb{P}(\bT^*_{m(j)} = \mathbf{u}_j^*) = \tfrac{1}{2},
\]
where $\mathbf{v}_j^* = (\mathrm{unif}(S_v, s), \mathrm{rep}(0, q-s))^\top$, $\mathbf{u}_j^* = (\mathrm{rep}(0, q-s), \mathrm{unif}(S_u, s))^\top$, $s = 0.2q$, and $\mathrm{unif}(S, s)$ denotes an $s$-vector of i.i.d. samples from the uniform distribution over $S$. We set $S_v = [-1, -0.5] \cup [0.5, 1]$ and $S_u = \{-1, 1\}$. This construction ensures both sparsity and heterogeneity at the row level, forming partially shared patterns across clients.

\smallskip
\noindent
\textbf{Methods compared.}
We compare four representative methods:
\begin{itemize}[leftmargin=2em, topsep=0pt]
\item[(a)] \textit{RowFed (proposed)}: Implements the proposed Sparse Row-wise Fusion (SROF) regularization within the federated system. It jointly performs row-level sparsity selection and variable-level clustering, effectively recovering the underlying heterogeneous structure.
\item[(b)] \textit{NonFed}: Fits each local model independently using a row-wise sparse penalty, enabling local sparsity recovery but lacking any cross-client communication or fusion.
\item[(c)] \textit{FedAvg}: Applies the global row-wise sparse penalty by aggregating all local data, ignoring client-specific heterogeneity. This often leads to model misspecification and poor estimation accuracy.
\item[(d)] \textit{PerFL-LSMA}: The personalized federated learning method based on sparse clustered association learning \citep{li2025personalized}, which imposes a matrix-wise fusion penalty. 
\end{itemize}
Expert PerFL-LSMA, tuning parameters are selected via a generalized information criterion (GIC) \citep{fan2013tuning}; we tuning PerFL-LSMA parameters as in {li2025personalized}.

\smallskip
\noindent
\textbf{Evaluation metrics.}
The estimation and prediction accuracies, as well as cluster identification quality, are evaluated through:
\begin{align*}
& \text{MSE-Est} = \frac{1}{M} \sum_{m=1}^{M} \frac{\|\widehat{\bT}_m - \bT^*_m\|_F^2}{pq}, \quad  \text{MSE-Pred} = \frac{1}{M} \sum_{m=1}^{M} \frac{\|\X_m (\widehat{\bT}_m - \bT^*_m)\|_F^2}{n_m q}, \\
& \text{RI} = \frac{\text{TP}_c + \text{TN}_c}{\text{TP}_c + \text{FP}_c + \text{FN}_c + \text{TN}_c},
\end{align*}
where MSE-Est measures estimation accuracy, MSE-Pred reflects predictive performance, and RI (Rand Index) quantifies the correctness of variable-level cluster identification. Here, $\text{TP}_c$, $\text{TN}_c$, $\text{FP}_c$, and $\text{FN}_c$ denote true positive, true negative, false positive, and false negative counts of pairwise clustering decisions.

Across heterogeneous settings, RowFed attains the best estimation/prediction errors and near-perfect cluster recovery, outperforming FedAvg and NonFed and matching PerFL-LSMA while offering clearer row-wise interpretability.

\smallskip
\noindent
\textbf{Estimation performance for heterogeneous model.}

\begin{figure}[h]
\centering
\begin{minipage}[t]{.48\linewidth}
  \centering
  \captionof{table}{Varied dimensions $(p,q)$. Means (s.d.) over 100 replicates. MSE-Est / MSE-Pred scaled by $10^2$.}
  \label{tab:hetero1}
  \resizebox{\linewidth}{!}{
  \begin{threeparttable}
  \begin{tabular}{lcccc}
  \toprule
  Method & $(p,q)$ & MSE-Est & MSE-Pred & RI$_c$ \\
  \midrule
        NonFed & (100,50) & 0.92(0.34) & 12.61(0.49) & --- \\
        FedAvg & (100,50) & 13.10(1.22) & 712.25(9.78) & --- \\
        PerFL-LSMA  & (100,50) & 0.62(0.06) & 6.21(0.28) & --- \\
        RowFed & (100,50) & \textbf{0.51(0.05)} & \textbf{5.07(0.17)} & \textbf{0.97} \\
        NonFed & (100,100) & 0.71(0.36) & 8.92(0.32) & --- \\
        FedAvg & (100,100) & 9.23(0.53) & 429.20(5.61) & --- \\
        PerFL-LSMA & (100,100) & \textbf{0.31(0.04)} & 5.70(0.57) & --- \\
        RowFed & (100,100) & 0.32(0.04) & \textbf{5.20(0.32)} & \textbf{0.98} \\
        NonFed & (300,100) & 0.33(0.10) & 12.15(0.24) & --- \\
        FedAvg & (300,100) & 6.21(0.41) & 626.01(4.12) & --- \\
        PerFL-LSMA & (300,100) & 0.18(0.06) & 6.21(0.69) & --- \\
        RowFed & (300,100) & \textbf{0.12(0.05)} & \textbf{4.10(0.31)} & \textbf{0.93} \\
  \bottomrule
  \end{tabular}
  \end{threeparttable}}
\end{minipage}
\hfill
\begin{minipage}[t]{.48\linewidth}
  \centering
  \captionof{table}{Varied federation sizes $(M,n_m)$. Means (s.d.) over 100 replicates. MSE-Est / MSE-Pred scaled by $10^2$.}
  \label{tab:hetero2}
  \resizebox{\linewidth}{!}{
  \begin{threeparttable}
  \begin{tabular}{lcccc}
  \toprule
  Method & $(M,n_m)$ & MSE-Est & MSE-Pred & RI$_c$ \\
  \midrule
        NonFed & (10,50) & 1.22(0.56) & 21.10(0.52) & --- \\
        FedAvg & (10,50) & 15.91(1.78) & 532.72(15.19) & --- \\
        PerFL-LSMA  & (10,50) & 0.77(0.34) & 10.10(0.46) & --- \\
        RowFed & (10,50) & \textbf{0.52(0.21)} & \textbf{7.21(0.41)} & \textbf{0.94} \\
        NonFed & (10,100) & 0.71(0.36) & 8.92(0.32) & --- \\
        FedAvg & (10,100) & 9.23(0.53) & 429.20(5.61) & --- \\
        PerFL-LSMA & (10,100) & \textbf{0.31(0.04)} & 5.70(0.57) & --- \\
        RowFed & (10,100) & 0.32(0.04) & \textbf{5.20(0.32)} & \textbf{0.98} 
        \\
        NonFed & (30,100) & 0.63(0.17) & 9.12(0.52) & --- \\
        FedAvg & (30,100) & 10.29(0.36) & 362.52(6.41) & --- \\
        PerFL-LSMA & (30,100) & 0.23(0.02) & 3.14(0.14) & --- \\
        RowFed & (30,100) & \textbf{0.14(0.05)} & \textbf{2.19(0.08)} & \textbf{0.96} \\
  \bottomrule
  \end{tabular}
  \end{threeparttable}}
\end{minipage}

\caption{Simulation results under heterogeneous settings. Left: varying $(p,q)$. Right: varying $(M,n_m)$.}
\end{figure}

\smallskip
\noindent
\textit{Varying dimensionality.}
Table~\ref{tab:hetero1} reports results for fixed federation size $(M,n_m)=(10,100)$ and varying $(p,q)$ over $100$ replications.
RowFed consistently attains the lowest MSE-Est/MSE-Pred and perfect (or near-perfect) RI, evidencing precise variable-level clustering.
FedAvg enforces a single global model and thus mis-specifies heterogeneous clients, yielding the largest errors.
NonFed estimates each client independently and recovers local sparsity, but it cannot pool cross-client signal, resulting in higher errors than RowFed.
PerFL-LSMA is competitive; however, RowFed achieves \emph{lower} errors while providing variable-level interpretability via row-wise fusion, in line with our modeling assumption.

\smallskip
\noindent
\textit{Varying federation size.}
Table~\ref{tab:hetero2} fixes $(p,q)=(100,100)$ and varies $(M,n_m)$.
As either the number of clients $M$ or the per-client sample size $n_m$ grows, RowFed’s estimation and prediction errors decrease markedly due to effective cross-client information sharing.
By contrast, NonFed shows no improvement with larger $M$—its accuracy is driven solely by $n_m$—and FedAvg remains consistently poor because it ignores personalization.
PerFL-LSMA remains strong across settings, yet RowFed delivers \emph{better} accuracy together with transparent variable-level clustering and interpretability.

%-------------------------------------------------
\subsection{Real Data Performance Comparison}

\smallskip
\noindent\textbf{Data and setup.}
We evaluate our methods on the \emph{Communities and Crime} dataset from the UCI repository, which integrates socio-economic variables (1990 Census), policing information (1990 LEMAS), and crime statistics (1995 FBI UCR) across 48 U.S. states. Following standard practice, all numeric variables are scaled to $[0,1]$, and missing values are imputed via $k$-NN. We take $q=28$ crime-related attributes as multivariate responses and $p=106$ socio-economic features as predictors. Each state is treated as a client; keeping states with at least three observations yields $M=45$ clients with per-state sample sizes ranging from $3$ to $279$ (Table~\ref{tab:state_sum}).

\begin{table}[h]
\centering
\caption{Summary of per-state sample sizes ($n_m$): five-number summary and mean.}
\label{tab:state_sum}
\begin{tabular}{ccccccc}
\toprule
 & Min. & Q1 & Median & Mean & Q3 & Max. \\
\midrule
$n_m$ & 3 & 20 & 31 & 49.16 & 48 & 279 \\
\bottomrule
\end{tabular}
\end{table}

\smallskip
\noindent\textbf{Methods and metric.}
Consistent with our modeling focus on variable-level structure, we compare the personalized, matrix-wise fusion baseline \textbf{PerFL-LSMA} with our row-wise fusion method \textbf{RowFed}. Performance is summarized by the per-state prediction mean squared error (MSE) and visualized via boxplots across the $45$ states.

\begin{figure}[h]
    \centering
    \includegraphics[width=0.5\textwidth, height=6cm]{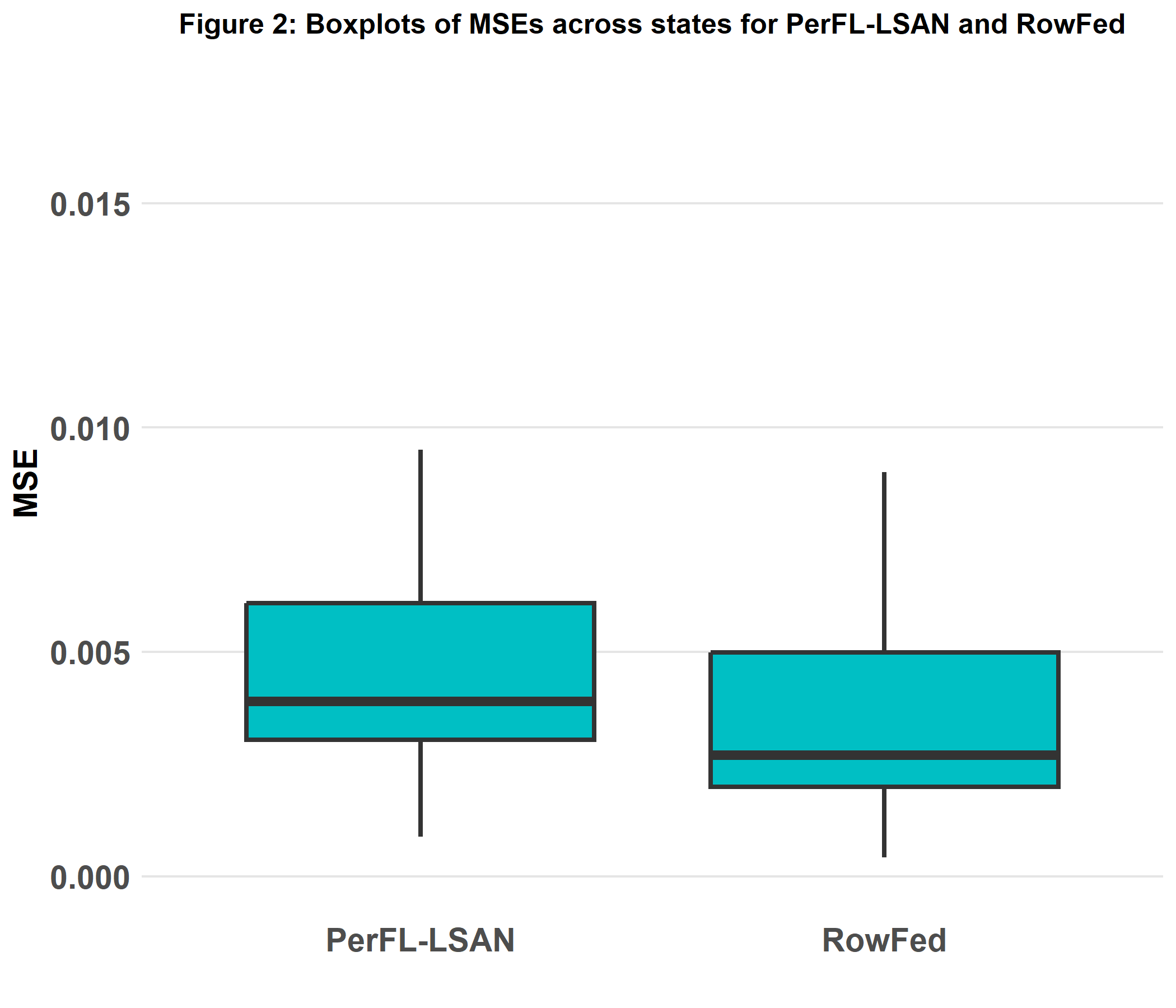}
    \caption{Boxplots of MSEs across states for PerFL-LSMA and RowFed.}
    \label{fig:boxplot}
\end{figure}

\smallskip
\noindent\textbf{Results.}
Figure~\ref{fig:boxplot} shows two boxplots (PerFL-LSMA vs.\ RowFed). RowFed achieves a lower median MSE and reduced dispersion across states, indicating more accurate and stable personalization. While PerFL-LSMA remains competitive, RowFed delivers \emph{better} predictive accuracy together with variable-level interpretability via row-wise fusion, aligning with the heterogeneous, row-structured effects anticipated in this application.

%-------------------------------------------------
\section{Conclusions}
We proposed \emph{Sparse Row-wise Fusion} (SROF) for personalized federated learning with multivariate responses and developed \emph{RowFed}, a communication-efficient linearized-ADMM algorithm with partial participation through a privacy-preserving procedure. 

By clustering variable-level effects across clients while inducing within-row sparsity, our framework bridges entry-wise and matrix-wise regimes, yielding interpretable models that respect heterogeneity and scale to large deployments. Theoretically, SROF attains an oracle property with asymptotic normality, and RowFed provably converges to a stationary solution; under random participation, the iterate gap contracts faster with higher participation probability. Empirically, across heterogeneous and homogeneous settings, RowFed reduces estimation/prediction error and strengthens variable-level cluster recovery relative to \textit{NonFed}, \textit{FedAvg}, and a personalized matrix-fusion baseline; a real-data study corroborates these gains. Future work includes generalized/nonlinear models, temporal dynamics, adaptive and private communication, and sharper guarantees under model misspecification and asynchrony.

%\bigskip

\newpage
\bibliographystyle{informs2014}
\bibliography{bib/bibliography}
\clearpage

%% Here starts the e-companion (EC)
%%%%%%%%%%%%%%%%%%%%%%%%%%%%%%%%%%%%%%%%%%%%%%%%%%%%%%%%%%
\ECSwitch

%\ECDisclaimer
%%%%%%%%%%%%%%%%%%%%%%%%%%%%%%%%%%%%%%%%%%%%%%%%%%%%%%%%%%

%%% Main head for the e-companion

\begin{APPENDICES}
% \newpage
% \section{Collection of Proofs}
% \input{AppendixProofs}

%	\numberwithin{equation}{section}
%	 \numberwithin{theorem}{section}
% \numberwithin{lemma}{section}
% \numberwithin{proposition}{section}

\begin{center}
\large{Online Appendix for "Row-wise Fusion Regularization:"}
\end{center}

\input{appendix/appendix}

\vspace{.2in}

\end{APPENDICES}

\end{document}

%% file: appendix/appendix.tex
% \setcounter{page}{1}
% \setcounter{section}{0}
% \renewcommand{\theequation}{A.\arabic{equation}}
% \setcounter{equation}{0}
% \renewcommand\thesection{\Alph{section}}
% \renewcommand\thesubsection{\Alph{section}.\arabic{subsection}}

% \begin{center}{\bf \Large Supplementary Material to "Row-wise Fusion Regularization: An Interpretable Personalized Federated Learning Framework in Large-Scale Scenarios"}

% \bigskip

% \end{center}

This Supplementary Material provides the additional technical details and all the proofs of theorems. The increased intuitions are as follow: $\X=(\X_1,\dots,\X_n)$, Vec($\X$) means recombine matrix $\X$ into vector $(\X_1\t,\dots,\X_n\t)\t$ 

%----------------------------------------------------
\section{Proof strecth of "SROF" Oracle}\label{sec:proSROF}
$\bT=\bf{W} \boldsymbol{\bX} $, where  $\boldsymbol{\bX}=\left(\bX_{(1)}, \ldots, \bX_{(S)}\right)\t$ and  $\bX_{(s^{\prime})}=\bX_{s}^{j}$  for  $s^{\prime}=\sum_{m=0}^{j-1} S_{m}+s$ .  $|\mathcal{G}|=\left\{\left|\mathcal{G}_{s}^{j}\right|, j \in\{1, \ldots, p\}, s \in\left\{1, \ldots, S_{j}\right\}\right\}=\left\{\left|\mathcal{G}_{1}\right|, \ldots,\left|\mathcal{G}_{S}\right|\right\}$ , where  $\left|\mathcal{G}_{s}^{j}\right|$  denotes the number of elements in  $\mathcal{G}_{s}^{j}$ .  $\left|\mathcal{G}_{\text {min}}\right|$  and  $\left|\mathcal{G}_{\text {max}}\right|$  are the true minimum and maximum unit sizes among all the groups.

Let  $\mathbb{X}_{1}$  and  $\mathbb{X}_{2}$  be  $n \times s_{k}$  and  $n \times(S-s_{k})$  submatrices of  $\mathbb{X}$  corresponding to the decomposition of  $\boldsymbol{\bX}=\left(\boldsymbol{\bX}_{\mathcal{K} 1}^{\top}, \boldsymbol{\bX}_{\mathcal{K} 2}^{\top}\right)^{\top}$. Similarly, let  $\boldsymbol{N}=\left(\boldsymbol{N}_{\mathcal{K} 1}^{\top}, \boldsymbol{N}_{\mathcal{K} 2}^{\top}\right)^{\top},|\mathcal{G}|=\left(\left|\mathcal{G}_{\mathcal{K} 1}\right|^{\top},\left|\mathcal{G}_{\mathcal{K} 2}\right|^{\top}\right)^{\top}$. Let  $\widehat{\boldsymbol{\bX}}, \widehat{\bT}$  denote the estimates of  $\bX, \bT$.

Denote  $d_{n}=\frac{1}{2} \min _{1 \leq j \leq q}\left\|\boldsymbol{\bX}_{0 \mathcal{K} 1,(j)}\right\|$  as the half of the minimum signal. 
% For a given vector  $\boldsymbol{b}=\left(b_{1}, \ldots, b_{t}\right)^{\top} \in \mathbb{R}^{t}$  and a symmetric matrix  $\boldsymbol{A}_{t \times t}$ , define  $\|\boldsymbol{b}\|_{\infty}=\max _{1 \leq w \leq t}\left|b_{w}\right|,\|\boldsymbol{A}\|=\|\boldsymbol{A}\|_{2}=\max _{\boldsymbol{b} \in \mathbb{R}^{t},\|\boldsymbol{b}\|=1}\|\boldsymbol{A} \boldsymbol{b}\|$ . Let  $\mathbb{E}_{\min }(\boldsymbol{A})\left(\mathbb{E}_{\text {max }}(\boldsymbol{A})\right)$  be the smallest (largest) eigenvalue of  $\boldsymbol{A}$ . For a matrix  $\boldsymbol{A}_{t \times w},\|\boldsymbol{A}\|_{2, \infty}=\max _{1 \leq i \leq t}\left\|\boldsymbol{A}_{i,},\right\|$ , where  $\left\|\boldsymbol{A}_{i},\right\|=\sqrt{\sum_{j=1}^{w} A_{i j}^{2}}$ . 
Set  $p_{\lambda}(0)=p_{\gamma}(0)=0, p_{\lambda}^{\prime}(a), p_{\gamma}^{\prime}(a)$  and  $p_{\lambda}^{\prime \prime}(a), p_{\gamma}^{\prime \prime}(a)$  denote the first and second derivations of penalty  $p_{\lambda}(a)$  and  $p_{\gamma}(a)$  about  a  respectively. Let  c  and  $c_{j}^{\prime}$  's denote some positive constants.

If the underlying integrative parameter, i.e., the matrix  $\bf{W}$  is known, we define an integration-oracle estimator by
\begin{align}\label{Ora}
\widetilde{\bX}=\underset{\bX \in \mathbb{R}^{S \times q}}{\operatorname{argmin}}\left\{\frac{1}{2 M}\|\boldsymbol{\Y}-\mathbb{X} \bX\|^{2}+\sum_{s=1}^{S}\left|\mathcal{G}_{s}\right| p_{\gamma}\left(\left\|\bX_{(s)}\right\|\right)\right\}    
\end{align}

Since the integrative relationship of the units, i.e., $\bf{W}$ is typically unknown in advance, the integration-oracle estimators are infeasible in practice, which however can shed light on the theoretical properties of the proposed estimators.

Theorem \ref{Xor} will prove that the oracle estimator can attain the local minimizer of \eqref{Ora} with probability tending to one. Hereafter we consider penalty functions $p_{\lambda}(\cdot)$ that satisfy the following conditions:
(C1) The penalty function $p_{\lambda}(t)$ is symmetric, nondecreasing and concave in $t \in[0,+\infty)$, and $p_{\lambda}(0)=0$. For $a>0$, $p_{\lambda}(t)$ is a constant for $t \geq a \lambda$. $p_{\lambda}^{\prime}(t)$ is increasing in $\lambda \in[0,+\infty)$. In addition, $p_{\lambda}^{\prime}(t)$ is nonincreasing in $t \in[0,+\infty)$ and $\lambda^{-1} p_{\lambda}^{\prime}(0+)=1$.

Next, for exploring the asymptotic properties of integration-oracle estimator $\widetilde{\bX}$, we first conclude the following theorem stating the sufficient conditions on the strict local minimizer of the penalized least square in \eqref{Ora}.

\begin{lem}[KKT Conditions]\label{KKT}
Assume that $p_{\gamma}(\cdot)$ satisfies Condition (C1). Then $\widetilde{\bX}=\left(\widetilde{\bX}_{\mathcal{K} 1}^{\top}, \widetilde{\bX}_{\mathcal{K} 2}^{\top}\right)^{\top}$ is a strict local minimizer of the penalized least square objective function defined in \eqref{Ora} if

\begin{subequations}
\begin{align}
&M^{-1} \mathbb{X}_{1}^{\top}\left(\Y-\mathbb{X}_{1} \widetilde{\bX}_{\mathcal{K} 1}\right)-\bbT_1(\widetilde{\bX}_{\mathcal{K} 1})\widetilde{\bX}_{\mathcal{K} 1}=0 \label{KKT1}\\
&M^{-1}\left\|\mathbb{X}_{2i}^{\top}\left(\Y-\mathbb{X}_{1} \widetilde{\bX}_{\mathcal{K} 1}\right)\right\|<\operatorname{diag}\left(\left|\mathcal{G}_{\mathcal{K} 2,i}\right|\right) p_{\gamma}^{\prime}(0+), i\in [S-s_k] \label{KKT2}\\
&M^{-1} \varsigma_{\min }\left(\mathbb{X}_{1}^{\top} \mathbb{X}_{1}\right)>\varsigma_{\max }(\bbT_2(\widetilde{\bX}_{\mathcal{K} 1}))\label{KKT3}
\end{align}    
\end{subequations}
\end{lem}

where $\bbT_1(\bX) = \operatorname{diag}\left(\frac{\left|\mathcal{G}_{s}\right|p_{\gamma}^{\prime}(\|\bX_{(s)}\|)}{\|\bX_{(s)}\|}, s \in [s_{row}] \right)$, $s_{row}$ is the number of rows in matrix $\bX$; $\bbT_2(\bX) = \bX\t \otimes \I_{s_{row}} \cdot \operatorname{diag}\left( \bH \right) \cdot \bX \otimes \I_{s_{row}} - \I_p \otimes \bbT_1(\bX)$, $\bH =\operatorname{diag}\left(\left|\mathcal{G}_{s}\right|\left(\frac{p_{\gamma}^{\prime\prime}(\|\bX_{(s)}\|)}{\|\bX_{(s)}\|^2}-\frac{p_{\gamma}^{\prime}(\|\bX_{(s)}\|)}{\|\bX_{(s)}\|^3}\right), s \in [s_{row}] \right)$

Here, \eqref{KKT1} and \eqref{KKT3} ensure that $\widetilde{\bX}$ is a strict local minimizer of \eqref{Ora} when constrained on the subspace of $\left\{\bX \in \mathbb{R}^{S}: \bX^{c}=0\right\}$, where $\bX^{c}$ denotes the submatrix of $\bX$ formed by the components outside the support of the non-zero rows of $\widetilde{\bX}$. Condition \eqref{KKT2} guarantees that the sparse estimator $\widetilde{\bX}$ is actually a strict local minimizer of \eqref{Ora} on the whole space $\mathbb{R}^{S}$.

To study the oracle property of the estimator \ref{obj}, we require the following conditions.
(C2) The noise vector $\bE_{i}$($\bE=(\bE_{1},\dots,\bE_{q})$) has sub-Gaussian tails such that $\operatorname{Pr}\left(\left|\boldsymbol{a}^{\top} \bE_{i}\right|<\|\boldsymbol{a}\| x\right) \geq 1-2 \exp \left(-c_{1} x^{2}\right)$ for any vector $\boldsymbol{a} \in \mathbb{R}^{n}$ and $x>0$, and $\mathrm{E}(\bE_{ij})^4 < \infty$ for $i \in [n], j \in [q]$. 

(C3) (i) $p_{\gamma}^{\prime}\left(d_{n}\right)=O\left(\sqrt{N_{\min}q} /\left(M\left|\mathcal{G}_{\text{max}}\right|\right)\right)$ for $d_{n} \gg \sqrt{s_k q / N_{\text{min}}}$; 
(ii) $\ln (S)=O\left(N_{\text{min}}^{\alpha}\right)$, $\alpha \in(0,1 / 2)$, where $S=\sum_{j=1}^{p} S_{j}$; 
(iii) For $\bPs \in \mathcal{N}_{0}$, where $\mathcal{N}_{0}=\left\{\bPs \in \mathbb{R}^{s_{k} \times q}:\left\|\bPs-\boldsymbol{\Phi}_{0 \mathcal{K} 1}\right\| \leq d_{n}\right\}$, $\varsigma_{\max }(\bbT_{2}(\bPs))=o\left(N_{\min} /M\right)$; 
(iv) $p_{\gamma}^{\prime}(0+) \gg \max \left\{ n\sqrt{s_{k} q / N_{\min}} /(M\left|\mathcal{G}_{\text{min}}\right|), n\sqrt{q \ln \left(N_{\min}\right) / N_{\min}^{1-\alpha}} / \left(M\mathcal{G}_{\text{min}}\right) \right\}$; (v) $N_{\min}=O\left(N_{\max}\right)$.

(C4) (i) $\mathbb{E}_{\min}\left(\mathbb{X}_{1}^{\top} \mathbb{X}_{1}\right) \geq c_{2} N_{\min}$; 
(ii) $\left\|\mathbb{X}_{2}^{\top} \mathbb{X}_{1}\right\|_{2, \infty}=O(n)$; 
(iii) $\left\|\mathbb{X}_{2 i}\right\|^{2}=N_{\mathcal{K} 2, i}$, $\sup _{i}\left\|\mathbb{X}_{i 1}\right\|^{2} \leq c_{3} q$, where $\mathbb{X}_{2 i}$ is the $i$th column of $\mathbb{X}_{2}$ and $\mathbb{X}_{i 1}$ is the $i$th row of $\mathbb{X}_{1}$; 
(iv) $\mathbb{E}_{\max}\left(\boldsymbol{X}_{k 1}^{\top} \boldsymbol{X}_{k 1}\right) \leq c_{4} n_{k}$, $k \in\{1, \ldots, K\}$, where $\boldsymbol{X}_{k 1}$ is the submatrix of $\boldsymbol{X}_{k}$ corresponding to the nonzero elements $\bT_{k 1}$ of $\bT$.

(C5) (i) $\lambda_2 \ge \left(2 \lambda_1\sqrt{q}+ \sqrt{ 1/c_{1}} \frac{\sqrt{s_kq\ln (s_kqn)}}{M} +  \tau \lambda_1\left|\mathcal{G}_{\min }\right|\right) /\left|\mathcal{G}_{\min }\right|$; 
(ii) $N_{\text{min}} \gg n^{3 / 4} q^{1 / 2}$.

The penalty MCP  used in this study satisfies (C1). Note that the assumption about sub-Gaussian tails of the noise vector in Condition (C2) is common in high-dimensional regression. The assumptions in (C3) (i) and (iv) give penalized level on nonzero and zero components respectively. Condition (C3) (ii) shows that if $S_{j}<c$, where $c$ is a constant, we have $p \leq S \leq c p$ and conclude NP-dimensional condition $\ln (p)=O\left(N_{\min}^{\alpha}\right)$, $\alpha \in(0,1 / 2)$. The combination of Conditions (C3) (iii) and (C4) (i) guarantees the sufficient condition (21) in Theorem 4.2. Condition (C4) (ii) is also used in [5]. Condition (C5) (ii) means $q=o\left(n^{1 / 3}\right)$.

\begin{thm}\label{Xor}
If Conditions (C1)-(C4) hold, there is a strict local minimizer $\widetilde{\bX}=\left(\widetilde{\bX}_{\mathcal{K} 1}^{\top}, \widetilde{\bX}_{\mathcal{K} 2}^{\top}\right)^{\top}$ such that

$$
\widetilde{\bX}_{\mathcal{K} 2}=0, \quad\left\|\widetilde{\bX}_{\mathcal{K} 1}-\bX_{0 \mathcal{K} 1}\right\|=O_{p}\left(\sqrt{s_kq / N_{\min}}\right)
$$

with probability tending to one as $N_{\min} \rightarrow \infty$.

\end{thm}

Theorem \ref{Xor} shows that, with high probability, there exists a consistent PHHS-regression estimator of $\widetilde{\bX}_{\mathcal{K}}$ by choosing a proper $\gamma$ which satisfies Condition (C1), (C3) (i) and (C3) (iv). Moreover, if $\min _{1 \leq j \leq q}\left\|\bX_{0 \mathcal{K} 1 j}\right\|>a \gamma$ under such SCAD and MCP penalties, $\widetilde{\bX}_{\mathcal{K} 1}=\widetilde{\bX}_{\mathcal{K} 1}^{*}$, that is, $\widetilde{\bX}$ just equals the oracle estimator.

\begin{thm}\label{Xnor}
Suppose the conditions in Theorem \ref{Xor} and (C5) (ii) holds. For any vector $a_n \in \mathbb{R}^{s_k}$, $c_n \in \mathbb{R}^{q}$, with $\|a_n\| = \|c_n\| =1$, as $n \to \infty$, we have
$$
\sigma_{n}\left(a_{n}\right)^{-1} a_{n}^{\top}\left(\widetilde{\bX}_{\mathcal{K} 1}-\bX_{0 \mathcal{K} 1}\right)c_n \rightarrow_{D} \mathcal{N}(0,1)
$$
and
$$
\sigma_{n}\left(a_{n}\right)=\sigma\left\{a_{n}^{\top}\left(\mathbb{X}_{1}^{\top} \mathbb{X}_{1}\right)^{-1} a_{n}\right\}^{1 / 2}.
$$
\end{thm}

We can easily obtain Theorem~\ref{Xnornew} based on Theorems~\ref{TXor} and~\ref{Xnor}.

\section{Proof of Theorem 1}%\ref{thm:1}}

\subsection{Proof of Proposition \ref{LCP}}

\begin{prop}[Lipschitz continuous property]\label{LCP}
\begin{align}
\left\|\nabla f(\bT)-\nabla f\left(\bT^{\prime}\right)\right\|_{2} & \leq \max \left\|\mathbf{x}_{m i}\right\|_{2}^{2} \cdot\left\|\bT-\bT^{\prime}\right\|_{2}
\end{align}
\end{prop}

For  $\forall \bT_{m}, \bT_{m}^{\prime} \in \mathbb{R}^{p}$ , we have
$$
\begin{aligned}
\left\|\nabla f\left(\bT_{m}\right)-\nabla f\left(\bT_{m}^{\prime}\right)\right\| 
& = \left\| \frac{1}{M n_m} \sum_{i=1}^{n_{m}}\left[ \x_{m i}(\bT_m\t \x_{m i} - \y_{m i})\t -\x_{m i}(\bT_m^{\prime\top} \x_{m i} - \y_{m i})\t\right] \right\| \\
& =\left\|\frac{1}{M n_{m}} \sum_{i=1}^{n_{m}}\left[\x_{m i}(\bT_m\t \x_{m i} - \bT_m^{\prime\top} \x_{m i})\t\right] \right\| \\
& =\frac{1}{M n_{m}} \left\| \sum_{i=1}^{n_{m}} \x_{m i}\x_{m i}\t(\bT_m - \bT_m^{\prime}) \right\| \\
& \leq \frac{1}{M n_{m}}\left\|\sum_{i=1}^{n_{m}} \x_{m i}\x_{m i}\t\right\|\left\| \bT_m - \bT_m^{\prime}\right\| \\
& \leq \frac{1}{M n_{m}} \sum_{i=1}^{n_{m}}\left\|\mathbf{x}_{m i}\right\|_{2}^{2} \cdot\left\|\bT_{m}-\bT_{m}^{\prime}\right\|\\
& \leq \frac{1}{M} \max _{i}\left\|\mathbf{x}_{m i}\right\|_{2}^{2} \cdot\left\|\bT_{m}-\bT_{m}^{\prime}\right\|
\end{aligned}
$$

It follows that, for  $\forall \bT, \bT^{\prime} \in \mathbb{R}^{p M}$ ,
$$
\begin{aligned}
\left\|\nabla f(\bT)-\nabla f\left(\bT^{\prime}\right)\right\|_{2} & =\left(\sum_{m=1}^{M}\left\|\nabla f_{\sigma}\left(\bT_{m}\right)-\nabla f_{\sigma}\left(\bT_{m}^{\prime}\right)\right\|^{2}\right)^{1 / 2} \\
& \leq \max \left\|\mathbf{x}_{m i}\right\|_{2}^{2} \cdot\left\|\bT-\bT^{\prime}\right\|
\end{aligned}
$$

Thus, the loss  $f(\bT)$  is Lipschitz continuous with  $\mu=\max \left\|\mathbf{x}_{m i}\right\|^{2}$ .

Next, we prove Lemma \ref{LRUT} using Proposition \ref{LCP}.
\subsection{Proof of lemma \ref{LRUT}}

\begin{lem}[Lagrangian Reduction of Updating  $\bT$]\label{LRUT}
Let Assumption 1 hold and  $\{\bT^{(t)}, \bf{\Phi}\mm,
 \bG\mm\}_{t=1}^{\infty}$  be generated by Algorithm 1. Let  $r>\rho \tau \cdot \varsigma_{\max }\left(\mathbf{A}^{\top} \mathbf{A}\right)+   \max (\tau \mu / 2,1)$ , then,
$$
\begin{array}{l}
\mathcal{L}_{\rho}\left(\bT^{(t+1)}, \bf{\Phi}\mm, \bG\mm\right)-\mathcal{L}_{\rho}\left(\bT^{(t)}, \bf{\Phi}\mm, \bG\mm\right) \\
\quad \leq-\left[\frac{\varsigma_{\min }(\mathbf{H})}{\tau}+\frac{\rho \cdot \varsigma_{\min }\left(\mathbf{A}^{\top} \mathbf{A}\right)}{2}-\frac{\mu}{2}\right]\left\|\bT^{(t+1)}-\bT^{(t)}\right\|_{2}^{2}
\end{array}
$$
\end{lem}

By the definition of  $\mathbf{A}$ , we have  $\varsigma_{\max }\left(\mathbf{A}^{\top} \mathbf{A}\right)=M+   1, \varsigma_{\min }\left(\mathbf{A}^{\top} \mathbf{A}\right)=1$ . The specification of  $r$  results in that  $\varsigma_{\min }(\mathbf{H}) / \tau+\rho \cdot \varsigma_{\min }\left(\mathbf{A}^{\top} \mathbf{A}\right) / 2-\mu / 2>0$ .

% Since  $\mathbf{A}$  is not full row rank, the upper bound of  $\| \bG^{(t+1)}-   \bG\mm \|_{2}^{2}$  cannot be given such as in Li and Pong (2015), Huang, Chen, and Huang (2019), and Lu et al. (2021). Instead, we show the convergence through the boundness of parameters  $\left\{\bG\mm\right\}_{t=1}^{\infty}$  and Lagrangian  $\left\{\mathcal{L}_{\rho^{(t)}}\left(\bT^{(t)}, \bf{\Phi}\mm, \bG\mm\right)\right\}_{t=1}^{\infty}$  during the updating process. Then, by the Bolzano-Weierstrass theorem, the sequence  $\left\{\bT^{(t)}, \bf{\Phi}\mm, \bG\mm\right\}_{t=1}^{\infty}$  must have a convergent subsequence. Therefore, if  $\mathcal{L}_{\rho}(\bT, \bP, \bG)$  is convex, the sequence  $\left\{\bT^{(t)}, \bf{\Phi}\mm, \bG\mm\right\}_{t=1}^{\infty}$  converges to the unique optimizer. However, in this work, we have a nonconvex optimization problem and we show that there exists a subsequence of  $\left\{\bT^{(t)}, \bf{\Phi}\mm, \bG\mm\right\}_{t=1}^{\infty}$  that satisfies the KKT conditions for minimization of the augmented Lagrangian.

Since $\bT\mmm$ is the minimizer of linearized augmented Lagrangian, $\widetilde{\mathcal{L}}\left(\bT ; \bT\mm,\bP^{(t)}, \bG^{(t)}\right)$  for the chosen clients, so  $\nabla \widetilde{\mathcal{L}}\left(\bT_{m}^{(t+1)} ; \bT^{(t)}, \bP^{(t)}, \bG^{(t)}\right)=0$  if  $m \in S^{(t)}$ . If  $m \notin S^{(t)}$, $\bT_{m}^{(t+1)}=\bT_{m}^{(t)}$ . Thus, we have

$$
\begin{aligned}
0= & \left< \left(\bT^{(t)}-\bT^{(t+1)}\right), \left[\nabla f\left(\bT^{(t)}\right)-\tau^{-1} \mathbf{H}\left(\bT^{(t)}-\bT^{(t+1)}\right)+\mathbf{A}^{\top} \bG^{(t)}+\rho^{(t)} \mathbf{A}^{\top}\left(\mathbf{A} \bT^{(t+1)}-\bP^{(t)}\right)\right]\right> \\
\leq & f\left(\bT^{(t)}\right)-f\left(\bT^{(t+1)}\right) + \frac{\mu}{2}\left\|\bT^{(t)}-\bT^{(t+1)}\right\|^{2}-\frac{1}{\tau}\left\|\bT^{(t)}-\bT^{(t+1)}\right\|_{\mathbf{H}}^{2}+\left<\bG\mm, \left(\mathbf{A} \bT^{(t)}-\mathbf{A} \bT^{(t+1)}\right)\right> \\
& +\left<\rho\left(\mathbf{A} \bT^{(t)}-\mathbf{A} \bT^{(t+1)}\right), \left(\mathbf{A} \bT^{(t+1)}-\bP^{(t)}\right)\right> \\
\leq & f\left(\bT^{(t)}\right)-f\left(\bT^{(t+1)}\right)+\frac{\mu}{2}\left\|\bT^{(t)}-\bT^{(t+1)}\right\|^{2}-\frac{1}{\tau}\left\|\bT^{(t)}-\bT^{(t+1)}\right\|_{\mathbf{H}}^{2}+\left< \bG\mm, \left(\mathbf{A} \bT^{(t)}-\bP^{(t)}\right)\right> \\
& -\left<\bG\mm, \left(\mathbf{A} \bT^{(t+1)}-\bP^{(t)}\right)\right>+\frac{\rho}{2}\left\|\mathbf{A} \bT^{(t)}-\bP^{(t)}\right\|^{2}-\frac{\rho}{2}\left\|\mathbf{A} \bT^{(t+1)}-\bP^{(t)}\right\|^{2}-\frac{\rho}{2}\left\|\mathbf{A} \bT^{(t)}-\mathbf{A} \bT^{(t+1)}\right\|^{2} \\
= & \mathcal{L}_{\rho}\left(\bT^{(t+1)}, \bP^{(t)}, \bG^{(t)}\right)-\mathcal{L}_{\rho}\left(\bT^{(t)}, \bP^{(t)}, \bG^{(t)}\right)+\frac{\mu}{2}\left\|\bT^{(t)}-\bT^{(t+1)}\right\|^{2} \\
& -\frac{1}{\tau}\left\|\bT^{(t)}-\bT^{(t+1)}\right\|_{\mathbf{H}}^{2}-\frac{\rho}{2}\left\|\bT^{(t)}-\bT^{(t+1)}\right\|_{\mathbf{A}^{\top} \mathbf{A}}^{2} \\
\leq 
& \mathcal{L}_{\rho^{(t)}}\left(\bT^{(t)}, \bP^{(t)}, \bG^{(t)}\right)-\mathcal{L}_{\rho^{(t)}}\left(\bT^{(t+1)}, \bP^{(t)}, \bG^{(t)}\right)-\left[\frac{\varsigma_{\min }\left(\mathbf{H}\right)}{\tau}+\frac{\rho \varsigma_{\min }\left(\mathbf{A}^{\top} \mathbf{A}\right)}{2}-\frac{\mu}{2}\right]\left\|\bT^{(t+1)}-\bT^{(t)}\right\|^{2},
\end{aligned}
$$
where the second step follows from Assumption 1, the third step follows from  $\left<(\mathbf{A}-\mathbf{B}), (\mathbf{B}-   \mathbf{C})\right>=2^{-1}\left(\|\mathbf{A}-\mathbf{C}\|^{2}-\|\mathbf{A}-\mathbf{B}\|^{2}-\|\mathbf{B}-\mathbf{C}\|^{2}\right)$ . Note that by construction  $\mathbf{H}=r \mathbf{I}-\rho \tau \mathbf{A}^{\top} \mathbf{A}$  and then  $\varsigma_{\min }(\mathbf{H}) \geq r-\rho \tau \varsigma_{\max }\left(\mathbf{A}^{\top} \mathbf{A}\right)$ . Hence, let  $r>\rho \tau \cdot \varsigma_{\max }\left(\mathbf{A}^{\top} \mathbf{A}\right)+\tau \mu/2 $ , then we have  $\varsigma_{\text {min }}(\mathbf{H})>\tau \mu/2$  and make sure that  $\tau^{-1} \varsigma_{\min }(\mathbf{H})+\rho \cdot \varsigma_{\min }\left(\mathbf{A}^{\top} \mathbf{A}\right) / 2-\mu/2 >0$ . Combining the conditions on  $r$ , we requires  $r>\rho \tau \cdot \varsigma_{\max }\left(\mathbf{A}^{\top} \mathbf{A}\right)+\max (\tau \mu/2,1)$ .

\subsection{Proof of Lemma \ref{Thlem2}}
In the following lemma, we construct boundness of sequences  $\left\{\bG\mm\right\}_{t=1}^{\infty}$  and  $\left\{\mathcal{L}^{(t)}\right\}_{t=1}^{\infty}$.

\begin{lem}\label{Thlem2}
Let Assumptions 1 -2 hold and  $\left\{\bT^{(t)}, \bf{\Phi}\mm, \bG\mm\right\}_{t=1}^{\infty}$  be a sequence generated by Algorithm 1. If  $\rho^{(t+1)}=\alpha \rho^{(t)}, \alpha>1 , then,  \left\{\bG\mm\right\}_{t=1}^{\infty}$  and  $\left\{\mathcal{L}^{(t)}\right\}_{t=1}^{\infty}$  are bounded.
\end{lem} 

%Proof of Lemma \ref{Thlem2}. 
The first-order optimality conditions for the updates (5) follows:
$0 \in \partial h_{\lambda}\left(\bP\mmm\right)-\bG\mm-\rho^{(t)}\left(\mathbf{A} \bT^{(t+1)}-\bP\mmm\right)$ .

Combining (S.1) and (4), we have  $\bG^{(t+1)} \in \partial h_{\lambda}\left(\bP\mmm\right)$ . Then, by Proposition S.1, we have  $\left\|\bG\mmm\right\|_{2}^{2} \leq c_{\lambda}$ , where  $c_{\lambda}$  is some finite constant decided by  $\boldsymbol{\lambda}$ . Hence,  $\left\{\bG\mm\right\}_{t=1}^{\infty}$  is bounded. Next, we prove that  $\left\{\mathcal{L}^{(t)}\right\}_{t=1}^{\infty}$  is bounded through the following steps.
1. Combining (4) and (2),
\begin{equation}\label{Lbound1}
\mathcal{L}^{(t+1)}-\mathcal{L}_{\rho^{(t)}}\left(\bT^{(t+1)}, \bP\mmm, \bG^{(t+1)}\right)=\frac{\rho^{(t+1)}-\rho^{(t)}}{2\left(\rho^{(t)}\right)^{2}}\left\|\bG\mmm-\bG\mm\right\|_{2}^{2}    
\end{equation}

2. Combining (4) and (2),
\begin{equation}\label{Lbound2}
 \mathcal{L}_{\rho^{(t)}}\left(\bT^{(t+1)}, \bP\mmm, \bG\mmm\right)-\mathcal{L}_{\rho^{(t)}}\left(\bT^{(t+1)}, \bP\mmm, \bG\mm\right)=\frac{1}{\rho^{(t)}}\left\|\bG\mmm-\bG\mm\right\|_{2}^{2}   
\end{equation}

3. By the updating rule of  $\bP\mmm$  in (5),
\begin{equation}\label{Lbound3}
\mathcal{L}_{\rho^{(t)}}\left(\bT^{(t+1)}, \bP\mmm, \bG\mm\right)-\mathcal{L}_{\rho^{(t)}}\left(\bT^{(t+1)}, \bf{\Phi}\mm, \bG\mm\right) \leq 0 .    
\end{equation}

4. By the updating of  $\bT^{(t+1)}$  in (6),
\begin{equation}\label{Lbound4}
\mathcal{L}_{\rho^{(t)}}\left(\bT^{(t+1)}, \bf{\Phi}\mm, \bG\mm\right)-\mathcal{L}^{(t)} \leq-\left[\frac{\varsigma_{\min }(\mathbf{H})}{\tau}+\frac{\rho^{(t)} \cdot \varsigma_{\min }\left(\mathbf{A}^{\top} \mathbf{A}\right)}{2}-\frac{\mu}{2}\right]\left\|\bT^{(t+1)}-\bT^{(t)}\right\|_{2}^{2}    
\end{equation}

Combining \eqref{Lbound1}-\eqref{Lbound4}, it follows
$$
\begin{aligned}
\mathcal{L}^{(t+1)}-\mathcal{L}^{(t)} & \leq \frac{\rho^{(t+1)}+\rho^{(t)}}{2\left(\rho^{(t)}\right)^{2}}\left\|\bG^{(t+1)}-\bG\mm\right\|_{2}^{2}-c^{(t)}\left\|\bT^{(t+1)}-\bT^{(t)}\right\|_{2}^{2} \\
& \leq \frac{\rho^{(t+1)}+\rho^{(t)}}{2\left(\rho^{(t)}\right)^{2}} c_{\boldsymbol{\lambda}}-c^{(t)}\left\|\bT^{(t+1)}-\bT^{(t)}\right\|_{2}^{2}
\end{aligned}
$$

Where  $c^{(t)}=\tau^{-1} S_{\min }(\mathbf{H})+\rho^{(t)} \cdot \varsigma_{\min }\left(\mathbf{A}^{\top} \mathbf{A}\right) / 2-\mu $ . Summing over  $t$  from $0$ to  $T$ , we have
\begin{equation}\label{Lbend0}
\mathcal{L}^{(t+1)}-\mathcal{L}^{(0)} \leq \frac{c_{\lambda}(\alpha+1)\left(1-\frac{1}{\alpha^{(t+1)}}\right)}{2 \rho^{(0)}\left(1-\frac{1}{\alpha}\right)}-\sum_{t=0}^{T} c^{(t)}\left\|\bT^{(t+1)}-\bT^{(t)}\right\|_{2}^{2}
\end{equation}

It follows that
\begin{equation}\label{Lbend}
\lim _{T \rightarrow \infty} \mathcal{L}^{(t+1)} \leq \mathcal{L}_{0}+\frac{c_{\lambda} \alpha(\alpha+1)}{2 \rho^{(0)}(\alpha-1)}<\infty    
\end{equation}

On the other side, since  $\rho^{T} \rightarrow \infty$ ,
$$
\lim _{T \rightarrow \infty} \mathcal{L}^{(t+1)} \geq \lim _{T \rightarrow \infty}-\frac{1}{2 \rho^{(t+1)}}\left\|\bG\mmm\right\|_{2}^{2}=0
$$

\subsection{Proof of Propsition \ref{Thpro}}
The next proposition follows directly from Lemma \ref{Thlem2}.
\begin{prop}\label{Thpro}
We have that  $\left\{\bf{\Phi}\mm\right\}_{l=1}^{\infty}$  is bounded if
$$
\frac{\min \left\{\lambda_{1}^{2}, \lambda_{2}^{2}\right\} \iota}{2}>\mathcal{L}_{0}+\frac{c_{\lambda} \alpha(\alpha+1)}{2 \rho^{(0)}(\alpha-1)}
$$
\end{prop} 

%Proof of Proposition \ref{Thpro}. 
By \eqref{Lbend}, we have
$$
\frac{\min \left\{\lambda_{1}^{2}, \lambda_{2}^{2}\right\} \iota}{2}>\lim _{t \rightarrow \infty} \mathcal{L}^{(t+1)}>\lim _{t \rightarrow \infty} h_{\lambda}\left(\bP\mmm\right)
$$
and hence,  $\left\{\bf{\Phi}\mm\right\}_{t=1}^{\infty}$  is bounded. Otherwise, if  $\left\|\bf{\Phi}\mm\right\|^{2} \rightarrow \infty$ , then for some  $d$, $\left\|\bP_{(d)}\right\| \rightarrow \infty$ , so  $p_{\lambda}\left(\left\|\bP_{(d)}\right\|\right) \geq \frac{\min \left\{\lambda_{1}^{2}, \lambda_{2}^{2}\right\} \iota}{2}$  and  $h_{\lambda}\left(\bP\mmm\right) \geq \frac{\min \left\{\lambda_{1}^{2}, \lambda_{2}^{2}\right\} \iota}{2}$ , contradiction.

\subsection{Details of proof for Theorem \ref{thm:1}}
Last, we can prove Theorem \ref{thm:1}. By Proposition \ref{Thpro},  $\left\{\bf{\Phi}\mm\right\}_{t=1}^{\infty}$  is bounded. Let  $\widetilde{\mathrm{X}}=\operatorname{diag}\left\{\mathbf{X}_{1}, \ldots, \mathbf{X}_{M}\right\}$ . By Lemma \ref{Thlem2},  $\left\{\mathcal{L}^{(t)}\right\}_{t=1}^{\infty}$  is bounded, and hence  $\left\{\widetilde{\mathbf{X}} \bT^{(t)}\right\}_{t=1}^{\infty}$  is bounded. By (4), as  $\rho^{(t)} \rightarrow \infty$ , we have
$$
\lim _{t \rightarrow \infty}\left\|\mathbf{A} \bT^{(t+1)}-\bP\mmm\right\|^{2}=\lim _{t \rightarrow \infty} \frac{1}{\rho^{(t)}}\left\|\bG\mmm-\bG\mm\right\|^{2}=0
$$

Since  $\left\{\bf{\Phi}\mm\right\}_{t=1}^{\infty}$  is bounded, we have  $\left\{\mathbf{A} \bT^{(t)}\right\}_{t=1}^{\infty}$  is bounded. Since  $\mathbf{A}$  is full column rank,  $\operatorname{ker}(\widetilde{\mathbf{X}}) \cap \operatorname{ker}(\mathbf{A})=\{0\}$ . Thus, we conclude that  $\left\{\bT^{(t)}\right\}_{l=1}^{\infty}$  is bounded. So  $\left\{\bT^{(t)}, \bf{\Phi}\mm, \bG\mm\right\}_{l=1}^{\infty}$  is bounded. Combining \eqref{Lbend0} and \eqref{Lbend},
$$
0 \leq \lim _{T \rightarrow \infty} \sum_{t=0}^{T} c^{(t)}\left\|\bT^{(t+1)}-\bT^{(t)}\right\|_{2}^{2} \leq \mathcal{L}^{(0)}+\frac{\lambda^{2} \alpha(\alpha+1)}{2 \rho^{(0)}(\alpha-1)}
$$

Since  $\lim _{T \rightarrow \infty} \sum_{t=0}^{T} c^{(t)}\left\|\bT^{(t+1)}-\bT\mm\right\|^{2}<\infty$ , we have
$$
\lim _{t \rightarrow \infty}\left\|\bT^{(t+1)}-\bT^{(t)}\right\|^{2}=0
$$

We also have, as $t \rightarrow \infty$
$$
\begin{aligned}
\left\|\bP\mmm-\bf{\Phi}\mm\right\|^{2} & =\left\|\left(\mathbf{A} \bT^{(t)}-\bf{\Phi}\mm\right)-\left(\mathbf{A} \bT^{(t+1)}-\bP\mmm\right)+\left(\mathbf{A} \bT^{(t+1)}-\mathbf{A} \bT^{(t)}\right)\right\|^{2} \\
& \leq\left\|\mathbf{A} \bT^{(t)}-\bf{\Phi}\mm\right\|^{2}+\left\|\mathbf{A} \bT^{(t+1)}-\bP\mmm\right\|^{2}+\left\| \mathbf{A} \bT^{(t+1)}-\mathbf{A} \bT^{(t)}\right\|^{2} \\
& \rightarrow 0
\end{aligned}
$$

Then, the sequence  $\left\{\bT\mm, \bP\mm, \bG\mm\right\}_{t=1}^{\infty}$  has a subsequence  $\left\{\bT^{t_{k}},\bP^{t_{k}}, \bG^{t_{k}}\right\}_{t_{k}=1}^{\infty}$ , which converges to a point  $\left\{\bT^{*},\bP^{*}, \bG^{*}\right\}$ , which satisfies the KKT conditions.

Through (4), we have  $\mathbf{A}^{*} \bT^{*}-\bP^{*}=\mathbf{0}$ . The first-order optimality conditions for the updates (3) - (4) follows:
$$
\left\{\begin{array}{rr}
0= & \nabla f\left(\bT^{(t)}\right)-r^{-1} \mathbf{H}\left(\bT^{(t)}-\bT^{(t+1)}\right)+\mathbf{A}^{\top} \bG\mm \\
& \quad+\rho^{(t)} \mathbf{A}^{\top}\left(\mathbf{A} \bT^{(t+1)}-\bf{\Phi}\mm\right) \\
0 \in & \partial h_{\lambda_{1}}\left(\bP\mmm\right)-\bG\mm-\rho^{(t)}\left(\mathbf{A} \bT^{(t+1)}-\bP\mmm\right)
\end{array}\right.
$$

Hence,
$$
\left\{\begin{array}{ll}
0= & \nabla f\left(\bT^{*}\right)+\mathbf{A}^{\top} \bG^{*} \\
0 \in & \partial h_{\lambda_{1}}\left(\bP^{*}\right)-\bG^{*}
\end{array}\right.
$$

Let  $\widehat{\bT}^{(t)}$  be the virtual update as if all the clients are selected. Note (S.6) is equivalent to
$$
\mathcal{L}^{(t+1)}-\mathcal{L}^{(t)} \leq \frac{\alpha+1}{2 \rho^{(t)}} c_{\boldsymbol{\lambda}}-c^{(t)} \sum_{m \in S^{(t)}}\left\|\bT_{m}^{(t+1)}-\bT_{m}^{(t)}\right\|^{2}
$$

By taking the expectation with respect to the sampling of clients, we have
$$
\mathbb{E}\left\{\mathcal{L}^{(t+1)}-\mathcal{L}^{(t)}\right\} \leq \frac{\alpha+1}{2 \rho^{(t)}} c_{\boldsymbol{\lambda}}-c^{(t)} r_{p}\left\|\widehat{\bT}^{(t+1)}-\bT^{(t)}\right\|^{2} \leq \frac{\alpha+1}{2 \rho^{(t)}} c_{\boldsymbol{\lambda}}-c^{(0)} r_{p}\left\|\bT^{(t+1)}-\bT^{(t)}\right\|^{2}
$$

where the second inequality follows from  $c^{(t)} \geq c^{(0)}$  and  $\left\|\widehat{\bT}^{(t+1)}-\bT^{(t)}\right\|_{2}^{2} \geq\left\|\bT^{(t+1)}-\bT^{(t)}\right\|^{2}$ . Summing over from  $t=0$  to  $T-1$  and rearranging, it follows that
$$
0 \leq \lim _{T \rightarrow \infty} \sum_{t=0}^{T}\left\|\bT^{(t+1)}-\bT^{(t)}\right\|^{2} \leq \frac{1}{r_{p}}\left[\frac{\mathcal{L}^{(0)}}{c^{(0)}}+\frac{\lambda^{2} \alpha(\alpha+1)}{2 c^{(0)} \rho^{(0)}(\alpha-1)}\right]
$$

\section{Proof of Theorem 4}%\ref{Xor}}

\subsection{Proof of Lemma \ref{KKT}}
We will first derive the necessary condition. In view of \ref{Ora}, we have

$$
\nabla (\frac{1}{2 M}\|\boldsymbol{\Y}-\mathbb{X} \bX\|^{2})=M^{-1}\left[\mathbb{X}^{T} \mathbf{Y}-\mathbb{X}^{T}\mathbb{X}\bX\right] \text { and } \nabla^{2} (\frac{1}{2 M}\|\boldsymbol{\Y}-\mathbb{X} \bX\|^{2})=-M^{-1} I \otimes \mathbb{X}^{T}\mathbb{X}
$$

It follows from the classical optimization theory that if $\widetilde{\bX}=\left(\widetilde{\bX}_{(1)}, \cdots, \widetilde{\bX}_{(S)}\right)^{T}$ is a local maximizer of the penalized likelihood (3), it satisfies the Karush-Kuhn-Tucker (KKT) conditions, i.e.

\begin{align}\label{KKTned}
    M^{-1} \mathbb{X}^{\top}\left(\Y-\mathbb{X}_{1} \widetilde{\bX}_{\mathcal{K} 1}\right)-\bf{V}=0
\end{align}

where  $\bf{V} = (\bf{V}_{\mathcal{K} 1}\t, \bf{V}_{\mathcal{K} 2}\t)\t$, $\bf{V}_{\mathcal{K} 1} \in \R^{s_{k} \times q} $, $\bf{V}_{\mathcal{K} 2} \in \R^{S-s_{k} \times q} $ . We have  $\bf{V}_{\mathcal{K} 1}=\bbT_1(\widetilde{\bX}_{\mathcal{K} 1})\widetilde{\bX}_{\mathcal{K} 1}$,  $\bbT_1(\bX) = \operatorname{diag}\left(\frac{\left|\mathcal{G}_{s}\right|p_{\gamma}^{\prime}(\|\bX_{(s)}\|)}{\|\bX_{(s)}\|}, s \in [s_{row}] \right)$, $s_{row}$ is the number of rows in matrix $\bX$; and $\|\bf{V}_{\mathcal{K} 2(i)}\| \in\left[-\left|\mathcal{G}_{\mathcal{K} 2,i}\right|\rho^{\prime}(0+), \left|\mathcal{G}_{\mathcal{K} 2,i}\right|\rho^{\prime}(0+)\right]$. 

It follows from the second order condition that

$$
M^{-1} \mathbb{E}_{\min }\left(\mathbb{X}_{1}^{\top} \mathbb{X}_{1}\right)>\mathbb{E}_{\max }(\bbT_2(\widetilde{\bX}_{\mathcal{K} 1}))
$$

where $\bbT_2(\bX) = \bX\t \otimes \I_{s_{row}} \cdot \operatorname{diag}\left( \bH \right) \cdot \bX \otimes \I_{s_{row}} - \I_p \otimes \bbT_1(\bX)$, $\bH =\operatorname{diag}\left(\left|\mathcal{G}_{s}\right|\left(\frac{p_{\gamma}^{\prime\prime}(\|\bX_{(s)}\|)}{\|\bX_{(s)}\|^2}-\frac{p_{\gamma}^{\prime}(\|\bX_{(s)}\|)}{\|\bX_{(s)}\|^3}\right), s \in [s_{row}] \right)$. It is easy to see that equation \eqref{KKTned} can be equivalently written as
$$
\begin{array}{l}
M^{-1} \mathbb{X}_{1}^{\top}\left(\Y-\mathbb{X}_{1} \widetilde{\bX}_{\mathcal{K} 1}\right)-\bbT_1(\widetilde{\bX}_{\mathcal{K} 1})\widetilde{\bX}_{\mathcal{K} 1}=0 \\
M^{-1}\left\|\mathbb{X}_{2i}^{\top}\left(\Y-\mathbb{X}_{1} \widetilde{\bX}_{\mathcal{K} 1}\right)\right\|<\operatorname{diag}\left(\left|\mathcal{G}_{\mathcal{K} 2,i}\right|\right) p_{\gamma}^{\prime}(0+), i\in [S-s_k] 
\end{array}
$$

where $\mathbb{X}_{2i}$ denotes the $i$ th column of $\mathbb{X}_2$ .

The remaining proof of sufficient condition is similar to that of \cite{Fan2009NPD}.

\subsection{Details of proof for Theorem \ref{Xor}}
To prove the consistency of integration-oracle estimators in Theorem \ref{Xor}, it suffices to show that under the given regularity conditions, there exists a strict local minimizer of 
$$
Q(\bX)=\frac{1}{2 M}\|\Y-\mathbb{X} \bX\|^{2}+\sum_{s=1}^{S}\left|\mathcal{G}_{s}\right| p_{\gamma}\left(\left|\bX_{(s)}\right|\right)
$$
such that (1) $\left\|\widetilde{\bX}_{\mathcal{K} 1}-\bX_{0 \mathcal{K} 1}\right\|=O_{p}\left(\sqrt{s_{k}q / N_{\min }}\right)$ (i.e., $\sqrt{s_{k}q / N_{\min}}$-consistency), and (2) $\widetilde{\bX}_{\mathcal{K} 2}=0$ with probability tending to 1.

Step 1: We first constrain $Q(\bX)$ on the $q$-dimensional subspace $\left\{\bX \in \mathbb{R}^{S}: \bX_{\mathcal{K} 2}=0\right\}$ of $\mathbb{R}^{S}$. This constrained penalized least square objective function is given by

$$
\bar{Q}(\bPs)=\frac{1}{2 M}\left\|\Y-\mathbb{X}_{1} \bPs\right\|^{2}+\sum_{s=1}^{q}\left|\mathcal{G}_{\mathcal{K} 1, s}\right| p_{\gamma}\left(\left|\bPs_{(s)}\right|\right)
$$

where $\bPs=\left(\bPs_{1}, \ldots, \bPs_{s_{k}}\right)^{\top}$. We then show that there exists a local minimizer $\tilde{\bX}_{\mathcal{K} 1}$ of $\bar{Q}(\bPs)$ such that $\left\|\widetilde{\bX}_{\mathcal{K} 1}-\bX_{0 \mathcal{K} 1}\right\|=O_{p}\left(\sqrt{s_kq / N_{\min }}\right)$. Next, we define an event

$$
A_{n}=\left\{\min _{\bPs \in \partial \mathcal{N}_{\tau}} \bar{Q}(\bPs)>\bar{Q}\left(\bX_{0 \mathcal{K} 1}\right)\right\}
$$

where $\partial \mathcal{N}_{\tau}$ denotes the boundary of the closed set $\mathcal{N}_{\tau}=\left\{\bPs \in \mathbb{R}^{s_{k}\times q}:\left\|\bPs-\bX_{0 \mathcal{K} 1}\right\| \leq \tau \sqrt{s_{k}q / N_{\min }}\right\}$ with $\tau>0$. The definition of $A_{n}$ implies that there exists a local minimizer $\widetilde{\bX}_{\mathcal{K} 1}$ of $\bar{Q}(\bPs)$ in $\mathcal{N}_{\tau}$. Thus we just show that $\operatorname{Pr}\left(A_{n}\right)$ is close to 1 as $N_{\min} \rightarrow \infty$ and $\tau$ is large. 

Condition (C3) (i) implies that for $\bPs \in \partial \mathcal{N}_{\tau}$, 
%we have $\min _{j}\left|\varphi_{j}\right| \geq d_{n}$, $\operatorname{sign}(\varphi)=\operatorname{sign}\left(\bX_{0 \mathcal{K} 1}\right)$ and 
$\left\|\bPs-\bX_{0 \mathcal{K} 1}\right\|_{2 \rightarrow \infty} \leq d_{n}$. We use Taylor's theorem for any $\bPs \in \partial \mathcal{N}_{\tau}$ and obtain

$$
\bar{Q}(\bPs)-\bar{Q}\left(\bX_{0 \mathcal{K} 1}\right)=\left<\bPs-\bX_{0 \mathcal{K} 1},-\bf{C}\right>+\operatorname{Vec}\left(\bPs-\bX_{0 \mathcal{K} 1}\right)^{\top} \mathbb{D}\operatorname{Vec}\left(\bPs-\bX_{0 \mathcal{K} 1}\right) / 2
$$

where $\bf{C}=\mathbb{X}_{1}^{\top} (\Y-\mathbb{X}_{1}\bX_{0 \mathcal{K} 1}) / M-\bbT_1(\widetilde{\bX}_{\mathcal{K} 1})\widetilde{\bX}_{\mathcal{K} 1}$, and $\mathbb{D}=\I \otimes \mathbb{X}_{1}^{\top} \mathbb{X}_{1} / M+ \bbT_2(\widetilde{\bX}_{\mathcal{K} 1}^*)$, and $\bX_{\mathcal{K} 1}^{*}$ lies on the line segment joining $\bPs$ and $\bX_{0 \mathcal{K} 1}$. Recall the definition of $\mathcal{N}_{0}$ in Condition (C3), we have $\bX_{\mathcal{K} 1}^{*} \in \mathcal{N}_{0}$, then we can combine Condition (C3) (iii) with (C4) (i) and obtain

$$
\mathbb{E}_{\min }(\mathbb{D}) \geq c_{2} N_{\min } / M-c_{2} N_{\min } /(2 M)=c_{2} N_{\min } /(2 M)
$$

For $\bPs \in \partial \mathcal{N}_{\tau}$, we have $\left\|\bPs-\bX_{0 \mathcal{K} 1}\right\| \leq \tau \sqrt{s_{k} q / N_{\min}}$ and then

$$
\bar{Q}\left(\bX_{0 K 1}\right)-\min _{\bPs \in \partial \mathcal{N}_{\tau}} \bar{Q}(\bPs) \leq \tau \sqrt{s_{k} q / N_{\min}}\left(\|\bC\|-\tau \sqrt{s_{k} q / N_{\min}} c_{2} N_{\min} /(2 M)\right)
$$

which along with Markov's inequality entails that

$$
\operatorname{Pr}\left(A_{n}\right) \geq \operatorname{Pr}\left(\|\bC\|^{2} \leq \tau^{2} s_{k} q c_{2}^{2} N_{\min} /\left(4 M^{2}\right)\right) \geq 1-\frac{4 M^{2} \mathrm{E}\left(\|\bC\|^{2}\right)}{\tau^{2} s_{k} q c_{2}^{2} N_{\min}}
$$

By Condition (C3) (i) and (C3) (v), we have

\begin{align*}
\mathrm{E}\left(\|\bC\|^{2}\right) & \leq \mathrm{E}\operatorname{tr}\left(\bE^{\top} \mathbb{X}_{1} \mathbb{X}_{1}^{\top} \bE\right) / M^{2}+s_{k}\left|\mathcal{G}_{\max}\right|^{2} p_{\gamma}^{\prime}\left(d_{n}\right)^{2} \\
& = q\sigma^{2} \operatorname{tr}\left(\mathbb{X}_{1}^{\top} \mathbb{X}_{1}\right) / M^{2}+q\left|\mathcal{G}_{\max}\right|^{2} p_{\gamma}^{\prime}\left(d_{n}\right)^{2} \\
& = O\left(s_{k} q N_{\min} / M^{2}\right)+O\left(s_{k} q N_{\min} / M^{2}\right)=O\left(s_{k} q N_{\min} / M^{2}\right)
\end{align*}

Thus $\operatorname{Pr}\left(A_{n}\right) \geq 1-O\left(1 / \tau^{2}\right)$.

Step 2: Now we generate a synthetic estimator $\widetilde{\bX}=\left(\widetilde{\bX}_{\mathcal{K} 1}^{\top}, \widetilde{\bX}_{\mathcal{K} 2}^{\top}\right)^{\top}$ with $\widetilde{\bX}_{\mathcal{K} 1}$ being the local minimizer of the constrained objective function $\bar{Q}(\bX)$ in (B.1) and $\widetilde{\bX}_{\mathcal{K} 2}=0$. Therefore, it suffices to check condition (20) of Theorem 4.2. Next, define an event

$$
B_{n}=\left\{\left\|\mathbb{X}_{2 i}^{\top} \bE\right\| / n<\left|\mathcal{G}_{\mathcal{K} 2, i}\right| p_{\gamma}^{\prime}(0+), i \in\{1, \ldots, S-s_k\}\right\}
$$

where $\mathbb{X}_{2 i}$ is the $i$th column of $\mathbb{X}_{2}$. Then under events $A_{n}, B_{n}$, Condition (C3) (iv) and (C4) (ii),

\begin{align*}
\left\|\mathbb{X}_{2 i}^{\top}\left(\Y-\mathbb{X}_{1} \widetilde{\bX}_{\mathcal{K} 1}\right)\right\| / M & \leq \left\|\mathbb{X}_{2 i}^{\top} \bE\right\| / M+M^{-1}\left\|\mathbb{X}_{2 i}^{\top} \mathbb{X}_{1}\right\|\left\|\widetilde{\bX}_{\mathcal{K} 1}-\bX_{0 \mathcal{K} 1}\right\| \\
& < \left|\mathcal{G}_{\mathcal{K} 2, i}\right| p_{\gamma}^{\prime}(0+)+O(1)\left\|\widetilde{\bX}_{\mathcal{K} 1}-\bX_{0 \mathcal{K} 1}\right\| \\
& = \left|\mathcal{G}_{\mathcal{K} 2, i}\right| p_{\gamma}^{\prime}(0+)(1+o(1))=\left|\mathcal{G}_{\mathcal{K} 2, i}\right| p_{\gamma}^{\prime}(0+)
\end{align*}

for sufficiently large $N_{\min}$. under Conditions (C2), we know that
\begin{align*}
\operatorname{Pr}\left(\left\|\boldsymbol{a}^{\top} \bE\right\|< q\|\boldsymbol{a}\| x\right)&\geq \sum_{i=1}^{q}\operatorname{Pr}\left( \left|\boldsymbol{a}^{\top} \bE_{i}\right|<\|\boldsymbol{a}\| x \right)\\
&\geq q\left(1-2 \exp \left(-c_{1} x^{2}\right)\right)\\
&\geq 1-2 \exp \left(-c_{1}q x^{2}\right)
\end{align*}

The last inequality is derived from $a(1-2\exp^{-b/a}) \ge 1 - 2\exp^{-b}$, when $a \ge 1$ and $b \ge 2a$. Hence, under (C4) (iii) and (C3) (iv) and the above equation, we have

\begin{align*}
\operatorname{Pr}\left\{M^{-1}\left\|\mathbb{X}_{2 i}^{\top} \bE\right\|<\left|\mathcal{G}_{\mathcal{K} 2, i}\right| p_{\gamma}^{\prime}(0+)\right\} & = \operatorname{Pr}\left\{\left\|\mathbb{X}_{2 i}^{\top} \bE\right\|<q\left\|\mathbb{X}_{2 i}\right\| \frac{M\left|\mathcal{G}_{\mathcal{K} 2, i}\right| p_{\gamma}^{\prime}(0+)}{q\left\|\mathbb{X}_{2 i}\right\|}\right\} \\
& \geq 1-2 \exp \left(-c_{1} \frac{M^{2}\left|\mathcal{G}_{\mathcal{K} 2, i}\right|^{2} p_{\gamma}^{\prime}(0+)^{2}}{q\left\|\mathbb{X}_{2 i}\right\|^{2}}\right) \\
& \geq 1-2 \exp \left(-c_{1} \frac{M^{2}\left|\mathcal{G}_{\text{min}}\right|^{2}}{qN_{\min}}\max \left( \frac{n^2 s_k q}{M^2 N_{\min} \left|\mathcal{G}_{\text{min}}\right|^2}, \frac{n^2 s_k \ln(N_{\min})}{N_{\min}^{(1-\alpha)}\left|\mathcal{G}_{\text{min}}\right|^2}\right)\right) \\
& \geq 1-2 \exp \left(-\max \left(\frac{n^2 s_k}{N_{\min}^2},  \frac{n^2\ln\left(N_{\min}\right) N_{\min}^{\alpha}}{N_{\min}^2}\right)\right) \\
& \geq 1-2 \exp \left(- \max \left(s_k, \ln \left(N_{\min}\right) N_{\min}^{\alpha}\right)\right) \\
& \geq 1-2 \exp \left(-N_{\min}^{\alpha} \ln \left(N_{\min}\right)\right)
\end{align*}

Hence,

\begin{align*}
    \operatorname{Pr}\left\{B_{n}\right\} &\geq \prod_{i=1}^{S-s_k}\left(1-2 \exp \left(-N_{\min }^{\alpha} \ln \left(N_{\min }\right)\right)\right) \geq 1-2(S-s_k) \exp \left(-N_{\min }^{\alpha} \ln \left(N_{\min }\right)\right) \\
    & \geq 1-2 S \exp \left(-N_{\min }^{\alpha} \ln \left(N_{\min }\right)\right)
\end{align*}

Then, by Condition (C3) (ii), we have

$$
\operatorname{Pr}\left(A_{n} \cap B_{n}\right)=1-\operatorname{Pr}\left(A_{n}^{c} \cup B_{n}^{c}\right) \geq 1-\operatorname{Pr}\left(A_{n}^{c}\right)-\operatorname{Pr}\left(B_{n}^{c}\right)=1-O\left(1 / \tau^{2}\right)
$$

where the last inequality holds when $N_{\min} \rightarrow \infty$.

\section{Proof of Theorem 2}%\ref{TXor}}

First, for a matrix $\boldsymbol{A}$, let $\|\boldsymbol{A}\|_{2\rightarrow\infty}$ denote the maximum absolute value of 2-norm row sum of $\boldsymbol{A}$, and we define  
$\ell_{p}(\bT)=\frac{1}{2M} \sum_{m=1}^{M} \frac{1}{n_{m}} \sum_{i=1}^{n_{m}} \|\y_{mi}-\bT^{\t}_m\x_{mi}\|^2 +\sum_{j=1}^{p} \sum_{m=1}^{M} p_{\lambda_{1}}\left(\|\bT_{m(j)}\|\right)
+\sum_{j=1}^{p} \sum_{m \leq m^{\prime}} p_{\lambda_{2}}\left(\|\bT_{m(j)}-\bT_{m^{\prime}(j)}\|\right)$. Let $T: \mathcal{M}_{\mathcal{G}} \rightarrow \mathbb{R}^{S}$ be the mapping such that $T(\bT)$ is the $S \times q$ matrix whose $s^{\prime}$th row equals the common value of $\bT_{m (j)}$ for $m \in \mathcal{G}_{s}^{j}, j \in\{1, \ldots, p\}, s \in\left\{1, \ldots, S_{j}\right\}$, where $s^{\prime}=\sum_{m=0}^{j-1} S_{m}+s$ and $S_{0}=0$. Let $T^{*}: \mathbb{R}^{Mp \times q} \rightarrow \mathbb{R}^{S \times q}$ be the mapping such that $T^{*}(\bT)=\left\{\left|\mathcal{G}_{s}^{j}\right|^{-1} \sum_{m \in \mathcal{G}_{s}^{j}} \bT_{m (j)}, j \in\{1, \ldots, p\}, s \in\left\{1, \ldots, S_{j}\right\}\right\}^{\top}$. Then through ranking the nonzero part of parameters ahead of zeros, the form of $\bT$ can be written as $\left(\bT_{\mathcal{K} 1}^{\top}, \bT_{\mathcal{K} 2}^{\top}\right)^{\top}$, and the true value is $\left(\bT_{0 \mathcal{K} 1}^{\top}, \mathbf{0}^{\top}\right)^{\top}$.

Consider the neighborhood of $\bT_{0}$:

$$
\Theta=\left\{\bT \in \mathbb{R}^{Mp \times q}: \bT_{\mathcal{K} 2}=0,\left\|\bT_{\mathcal{K} 1}-\bT_{0 \mathcal{K} 1}\right\|_{2\rightarrow\infty} \leq \tau\sqrt{q/N_{\min }}\right\}
$$

where $\tau$ is a constant and $\tau>0$. For any $\bT \in \mathbb{R}^{Mp \times q}$, let $\bT^{*}=T^{-1}\left(T^{*}(\bT)\right)$ and by the result in Theorem \ref{Xor}, there is an event $E_{1}$ such that in the event $E_{1}$,

$$
\widetilde{\bT}_{\mathcal{K} 2}=0,\left\|\widetilde{\bT}_{\mathcal{K} 1}-\bT_{0 \mathcal{K} 1}\right\|_{2 \rightarrow\infty} \leq \tau \sqrt{q/N_{\min }}
$$

and $\operatorname{Pr}\left(E_{1}^{C}\right) \leq O\left(1 / \tau^{2}\right)$. Hence $\widetilde{\bT} \in \Theta$ in $E_{1}$. There is an event $E_{2}$ such that $\operatorname{Pr}\left(E_{2}^{C}\right) \leq 2 M / n$. In $E_{1} \cap E_{2}$, there is a neighborhood of $\widetilde{\bT}$, denoted by $\Theta_{n}$, and $\operatorname{Pr}\left(E_{1} \cap E_{2}\right) \geq 1-\left(O\left(1 / \tau^{2}\right)+2 / n\right)$. We show that $\widetilde{\bT}$ is a local minimizer of the objective function with probability approaching 1 such that (1) $\ell_{p}(\bT^{*}) \geq \ell_{p}\left(\widetilde{\bT}\right)$ for any $\bT^{*}\in\Theta$ and $\bT^{*} \ne \widetilde{\bT}$.(2)$\ell_{p}(\bT) \geq \ell_{p}\left(\bT^{*}\right)$ for any $\bT \in \Theta_{n} \cap \Theta$ for sufficiently large $n$ and $N_{\min}$.

Step 1: Let $T^{*}(\boldsymbol{\bT})=\left(\bX_{1}, \ldots, \bX_{S}\right)^{\top}$. When $S_{j}=1$, for all $m \neq m^{\prime}$, $\bT_{m (j)}^{*}=\bT_{m^{\prime} (j)}^{*}$, hence $p_{\lambda}\left(\left\|\bT_{m (j)}^{*}-\bT_{m^{\prime} (j)}^{*}\right\|\right)=0$. Next, we show that $\left\|\bX_{s}^{j}-\bX_{s^{\prime}}^{j}\right\|>a \lambda$ for all $\left\{j: S_{j}>1, j \in\{1, \ldots, p\}\right\}, s \neq s^{\prime}$, so that $p_{\lambda}\left(T^{*}(\bT)\right)=C_{n}$ for any $\bT \in \Theta$, where $C_{n}$ is a constant which does not depend on $\bT$. When $S_{j}>1$,

$$
\left\|\bX_{s}^{j}-\bX_{s^{\prime}}^{j}\right\| \geq\left\|\bX_{0 s}^{j}-\bX_{0 s^{\prime}}^{j}\right\|-2\left\|\bX_{s}^{j}-\bX_{0 s}^{j}\right\|_{2\rightarrow\infty}
$$

And
\begin{align*}
\left\|\bX_{s}^{j}-\bX_{0 s}^{j}\right\|_{2\rightarrow\infty} & =\sup _{j} \sup _{s}\left\|\sum_{m \in \mathcal{G}_{s}^{j}} \bT_{m (j)} / \left| \mathcal{G}_{s}^{j}\right|-\bX_{0 s}^{j}\right\| \\
& = \sup _{j} \sup _{s}\left\|\sum_{m \in \mathcal{G}_{s}^{j}}\left(\bT_{m (j)}-\bT_{0 m (j)}\right)\right\| /\left|\mathcal{G}_{s}^{j}\right| \\
& \leq \sup _{j} \sup _{s} \sup _{m \in \mathcal{G}_{s}^{j}}\left\|\bT_{m (j)}-\bT_{0 m (j)}\right\| \\
& = \left\|\bT-\bT_{0}\right\|_{2\rightarrow\infty} \leq \tau \sqrt{q/N_{\min }} 
\end{align*}

Hence by $a \lambda \gg  \sqrt{q/N_{\min }}$ 

$$
\left\|\bX_{s}^{j}-\bX_{s^{\prime}}^{j}\right\|=b_{n}-2 \tau \sqrt{q/N_{\min }}>a \lambda
$$

By the result in Theorem \ref{Xor}, (1) is proved.

Step 2: For a positive sequence $t_{n}$, let 

$$
\Theta_{n}=\left\{\bT:\|\bT-\widetilde{\bT}\|_{2\rightarrow\infty} \leq t_{n}\right\}.
$$ 

For $\bT \in \Theta_{n} \cap \Theta$, by Taylor's expansion (only for $\bT_{0 m (j)} \neq \mathbf{0}$), we have

$$
\ell_{p}(\bT)-\ell_{p}\left(\bT^{*}\right)=\Gamma_{1}+\Gamma_{2}+\Gamma_{3}
$$

where
$$
\begin{array}{l}
\Gamma_{1}=-\frac{1}{M} \sum_{M=1}^{m}\frac{1}{n_m}\left< \boldsymbol{X}_{m1}\t(\Y_{m1}-\X_{m1}\bT_{m1}^t),\bT_{m1}-\bT_{m1}^*\right>, \\
\Gamma_{2}=\sum_{m=1}^{M}\sum_{j=1}^{p}  \left(\frac{\partial p_{\lambda_1}\left(\bT^{t}\right)}{\partial \bT_{m (j)}}\right)\t\left(\bT_{m (j)}-\bT_{m (j)}^{*}\right) I\left(\bT_{0 m (j)} \neq 0\right) \\
\Gamma_{3}=\sum_{m=1}^{M}\sum_{j=1}^{p}  \left(\frac{\partial p_{\lambda_2}\left(\bT^{t}\right)}{\partial \bT_{m (j)}}\right)\t\left(\bT_{m (j)}-\bT_{m (j)}^{*}\right) I\left(\bT_{0 m (j)} \neq 0\right)
\end{array}
$$
in which $\bT_{m}^{t}=t \bT_m+(1-t) \bT_m^{*}$ for some $t \in(0,1)$. Moreover,

\begin{align*}
\Gamma_{2} & \geq -\sum_{j=1}^{p} \sum_{m=1}^{M}\left\| \frac{\partial p_{\lambda_1}\left(\bT^{t}\right)}{\partial \bT_{m (j)}}\right\|\left\|\left(\bT_{m (j)}-\bT_{m (j)}^{*}\right) I\left(\bT_{0 m (j)} \neq 0\right)\right\| \\
& \geq-\sqrt{q}\lambda_1 \sum_{j=1}^{p} \sum_{m=1}^{M}\left\|\left(\bT_{m (j)}-\bT_{m (j)}^{*}\right) I\left(\bT_{0 m (j)} \neq 0\right)\right\| \\
& \geq -\lambda_1 \sum_{j=1}^{p} \sum_{s=1}^{S_{j}} \sum_{m \in \mathcal{G}_{s}^{j}} \sum_{m^{\prime} \in \mathcal{G}_{s}^{j}}\left\|\left(\bT_{m (j)}-\bT_{m^{\prime} (j)}\right)\right\| /\left|\mathcal{G}_{s}^{j}\right| I\left(\bT_{0 m (j)} \neq 0 \text { or } \bT_{0 m^{\prime} (j)} \neq 0\right) \\
& =-\sum_{j=1}^{p} \sum_{s=1}^{S_{j}} \sum_{\left\{m, m^{\prime} \in \mathcal{G}_{s}^{j}, m<m^{\prime}\right\}} \frac{2 \sqrt{q}\lambda_1}{\left|\mathcal{G}_{s}^{j}\right|}\left\|\bT_{m (j)}-\bT_{m^{\prime} (j)}\right\| I\left(\bT_{0 m (j)} \neq 0 \text { or } \bT_{0 m^{\prime} (j)} \neq 0\right)
\end{align*}

Furthermore,
\begin{align*}
\Gamma_{3}= & \sum_{j=1}^{p}\{\sum_{m^{\prime}>m} p_{\lambda_2}^{\prime}\left(\|\bT_{m (j)}^{t}-\bT_{m^{\prime} (j)}^{t}\|\right)\frac{(\bT_{m (j)}^{t}-\bT_{m^{\prime} (j)}^{t})}{\|\bT_{m (j)}^{t}-\bT_{m^{\prime} (j)}^{t}\|}\t\left(\bT_{m (j)}-\bT_{m (j)}^{*}\right) I\left(\bT_{0 m (j)} \neq 0\right)\\
& +\sum_{m^{\prime}<m} p_{\lambda_2}^{\prime}\left(\|\bT_{m (j)}^{t}-\bT_{m^{\prime} (j)}^{t}\|\right)\frac{(\bT_{m (j)}^{t}-\bT_{m^{\prime} (j)}^{t})}{\|\bT_{m (j)}^{t}-\bT_{m^{\prime} (j)}^{t}\|}\t\left(\bT_{m (j)}-\bT_{m (j)}^{*}\right) I\left(\bT_{0 m (j)} \neq 0\right)\} \\
= & \sum_{j=1}^{p}\{\sum_{m^{\prime}>m} p_{\lambda_2}^{\prime}\left(\|\bT_{m (j)}^{t}-\bT_{m^{\prime} (j)}^{t}\|\right)\frac{(\bT_{m (j)}^{t}-\bT_{m^{\prime} (j)}^{t})}{\|\bT_{m (j)}^{t}-\bT_{m^{\prime} (j)}^{t}\|}\t\left(\bT_{m (j)}-\bT_{m (j)}^{*}\right) I\left(\bT_{0 m (j)} \neq 0\right)\\
& +\sum_{m^{\prime}>m} \frac{p_{\lambda_2}^{\prime}\left(\|\bT_{m^{\prime} (j)}^{t}-\bT_{m (j)}^{t}\|\right)}{\|\bT_{m^{\prime} (j)}^{t}-\bT_{m (j)}^{t}\|}(\bT_{m^{\prime} (j)}^{t}-\bT_{m (j)}^{t})\t\left(\bT_{m^{\prime} (j)}-\bT_{m^{\prime} (j)}^{*}\right) I\left(\bT_{0 m^{\prime} (j)} \neq 0\right)\} \\
= & \sum_{j=1}^{p}\{\sum_{m^{\prime}>m} p_{\lambda_2}^{\prime}\left(\|\bT_{m (j)}^{t}-\bT_{m^{\prime} (j)}^{t}\|\right)\frac{(\bT_{m (j)}^{t}-\bT_{m^{\prime} (j)}^{t})}{\|\bT_{m (j)}^{t}-\bT_{m^{\prime} (j)}^{t}\|}\t\left(\bT_{m (j)}-\bT_{m (j)}^{*}\right) I\left(\bT_{0 m (j)} \neq 0\right)\\
& -\sum_{m^{\prime}>m} \frac{p_{\lambda_2}^{\prime}\left(\|\bT_{m^{\prime} (j)}^{t}-\bT_{m (j)}^{t}\|\right)}{\|\bT_{m^{\prime} (j)}^{t}-\bT_{m (j)}^{t}\|}(\bT_{m (j)}^{t}-\bT_{m^{\prime} (j)}^{t})\t\left(\bT_{m^{\prime} (j)}-\bT_{m^{\prime} (j)}^{*}\right) I\left(\bT_{0 m^{\prime} (j)} \neq 0\right)\} \\
= & \sum_{j=1}^{p}\{\sum_{m^{\prime}>m} p_{\lambda_2}^{\prime}\left(\|\bT_{m (j)}^{t}-\bT_{m^{\prime} (j)}^{t}\|\right)\frac{(\bT_{m (j)}^{t}-\bT_{m^{\prime} (j)}^{t})}{\|\bT_{m (j)}^{t}-\bT_{m^{\prime} (j)}^{t}\|}\t\\
&\left(\left(\bT_{m (j)}-\bT_{m (j)}^{*}\right) I\left(\bT_{0 m (j)} \neq 0\right)
-\left(\bT_{m^{\prime} (j)}-\bT_{m^{\prime} (j)}^{*}\right) I\left(\bT_{0 m^{\prime} (j)} \neq 0\right)\right)\} \\
= & \sum_{j=1}^{p} \sum_{s=1}^{S_{j}} \sum_{\left\{m, m^{\prime} \in \mathcal{G}_{s}^{j}, m<m^{\prime}\right\}} p_{\lambda_2}^{\prime}\left(\|\bT_{m (j)}^{t}-\bT_{m^{\prime} (j)}^{t}\|\right)\frac{(\bT_{m (j)}^{t}-\bT_{m^{\prime} (j)}^{t})}{\|\bT_{m (j)}^{t}-\bT_{m^{\prime} (j)}^{t}\|}\t\left(\bT_{m (j)}-\bT_{m^{\prime} (j)}\right)  \\
& I\left(\bT_{0 m (j)} \neq 0 \text { or } \bT_{0 m^{\prime} (j)} \neq 0\right)+\sum_{j=1}^{p} \sum_{s<s^{\prime}} \sum_{\left\{m \in \mathcal{G}_{s}^{j}, m^{\prime} \in \mathcal{G}_{s^{\prime}}^{j}\right\}} p_{\lambda_2}^{\prime}\left(\|\bT_{m (j)}^{t}-\bT_{m^{\prime} (j)}^{t}\|\right)\frac{(\bT_{m (j)}^{t}-\bT_{m^{\prime} (j)}^{t})}{\|\bT_{m (j)}^{t}-\bT_{m^{\prime} (j)}^{t}\|}\t\\
&\left(\left(\bT_{m (j)}-\bT_{m (j)}^{*}\right) I\left(\bT_{0 m (j)} \neq 0\right)
-\left(\bT_{m^{\prime} (j)}-\bT_{m^{\prime} (j)}^{*}\right) I\left(\bT_{0 m^{\prime} (j)} \neq 0\right)\right)
\end{align*}

where the last equality holds due to the fact that for $m, m^{\prime} \in \mathcal{G}_{s}^{j}$, $\bT_{m (j)}^{*}=\bT_{m^{\prime} (j)}^{*}$ and $\bT_{m (j)}^{t}-\bT_{m^{\prime} (j)}^{t}=t \bT_{m (j)}+(1-t) \bT_{m (j)}^{*}-t \bT_{m^{\prime} (j)}-(1-t) \bT_{m^{\prime} (j)}^{*}=t \bT_{m (j)}-t \bT_{m^{\prime} (j)}$. The proof in theorem \ref{Xor} means that $\left\|\bT_{m (j)}^{*}-\bT_{0 m (j)}\right\| \leq \tau \sqrt{q/N_{\min }}$. Combined with $\bT^{t}=t \bT+(1-t) \bT^{*}$ and $\left\|\bT_{m (j)}-\bT_{0 m (j)}\right\| \leq \tau\sqrt{q/N_{\min }}$, then

\begin{align*}
\sup _{j} \sup _{m}\left\|\bT_{m (j)}^{t}-\bT_{0 m (j)}\right\| & =\sup _{j} \sup _{m}\left\|t \bT_{m (j)}+(1-t) \bT_{m (j)}^{*}-\bT_{0 m (j)}\right\| \\
& = \sup _{j} \sup _{m}\left\|t\left(\bT_{m (j)}-\bT_{0 m (j)}\right)+(1-t)\left(\bT_{m (j)}^{*}-\bT_{0 m (j)}\right)\right\| \\
& \leq \sup _{j} \sup _{m} t\left\|\bT_{m (j)}-\bT_{0 m (j)}\right\|+(1-t)\left\|\bT_{m (j)}^{*}-\bT_{0 m (j)}\right\| \leq \tau\sqrt{q/N_{\min }}.
\end{align*}

For $s \neq s^{\prime}, m \in \mathcal{G}_{s}^{j}, m^{\prime} \in \mathcal{G}_{s^{\prime}}^{j}$ 

$$
\left\|\bT_{m j}^{t}-\bT_{m^{\prime} j}^{t}\right\| \geq \min _{m \in \mathcal{G}_{s}^{j}, m^{\prime} \in \mathcal{G}_{s^{\prime}}^{j}}\left\|\bT_{0 m (j)}-\bT_{0 m^{\prime} (j)}\right\|-2\left\|\bT_{m (j)}^{t}-\bT_{0 m (j)}\right\| \geq b_{n}-2 \tau\sqrt{q/N_{\min }}>a \lambda_2
$$

and thus $p_{\lambda_2}^{\prime}\left(\|\bT_{m (j)}^{t}-\bT_{m^{\prime} (j)}^{t}\|\right)=0$. Therefore,

\begin{align*}
\Gamma_{3}&=\sum_{j=1}^{p} \sum_{s=1}^{S_{j}} \sum_{\left\{m, m^{\prime} \in \mathcal{G}_{s}^{j}, m<m^{\prime}\right\}} p_{\lambda_2}^{\prime}\left(\|\bT_{m (j)}^{t}-\bT_{m^{\prime} (j)}^{t}\|\right)\left\|\bT_{m (j)}-\bT_{m^{\prime} (j)}\right\|I\left(\bT_{0 m (j)} \neq 0 \text { or } \bT_{0 m^{\prime} (j)} \neq 0\right)
\end{align*}

Furthermore, by the same reasoning as in the proof in the theorem \ref{Xor}, we have

$$
\left\|\bT^{*}-\widetilde{\bT}\right\|_{2\rightarrow\infty}=\|\bT-\widetilde{\bT}\|_{2\rightarrow\infty} \leq \tau \sqrt{q/N_{\min }}
$$

Then

\begin{align*}
\left\|\bT_{m (j)}^{t}-\bT_{m^{\prime} (j)}^{t}\right\| & \leq 2\sup_{m}\left\|\bT_{m (j)}^{t}-\bT_{m (j)}^{*}\right\|=2\sup_{m}\left\| t \bT_{m(j)}+(1-t)\bT_{m(j)}^{*}-\bT_{m(j)}^{*}\right\| \\
&\leq 2\sup_{m}\left\|\bT_{m(j)}-\bT_{m(j)}^{*}\right\|
\leq 2\sup_{m}\left(\|\bT_{m(j)}-\bT_{0m(j)}\|+\left\|\bT^{*}_{m(j)}-\bT_{0m(j)}\right\|\right) \leq 4 t_{n}
\end{align*}

Hence $p_{\lambda_2}^{\prime}\left(\left\|\bT_{m (j)}^{t}-\bT_{m^{\prime} (j)}^{t}\right\|\right) \geq p_{\lambda_2}^{\prime}\left(4 t_{n}\right)$ by Condition (C1). As a result,

$$
\Gamma_{3} \geq \sum_{j=1}^{p} \sum_{s=1}^{S_{j}} \sum_{\left\{m, m^{\prime} \in \mathcal{G}_{s}^{j}, m<m^{\prime}\right\}} p_{\lambda}^{\prime}\left(4 t_{n}\right)\left\|\bT_{m (j)}-\bT_{m^{\prime} (j)}\right\| I\left(\bT_{0 m (j)} \neq 0 \text { or } \bT_{0 m^{\prime} (j)} \neq 0\right)
$$

By Condition (C2), for $m \in\{1, \ldots, M\}$, and $\left\{j: \bT_{0 m (j)} \neq 0\right\}$,

$$
\operatorname{Pr}\left(\left\|\boldsymbol{X}_{m 1}^{\top} \bE_m\right\|>\sqrt{ 1/c_{1}} \sqrt{s_kq\ln (s_kqn)}\right) \leq\sum_{i,j}\operatorname{Pr}\left(\left\|X_{m 1 ,i}^{\top} \bE_{mj}\right\|>\sqrt{1 / c_{1}} \sqrt{\ln (s_kqn)} \right) \leq 2 / n
$$

Then, by Condition (C4) (iv),

\begin{align*}
\Gamma_{1} & =-\frac{1}{M} \sum_{m=1}^{M}\frac{1}{n_m}\left\|\bT_{m1}-\bT_{m1}^{*}\right\| \left\|\boldsymbol{X}_{m1}\t(\Y_{m1}-\X_{m1}\bT_{m1}^t)\right\| \\
&=-\frac{1}{M} \sum_{m=1}^{M}\frac{1}{n_m}\left\|\bT_{m1}-\bT_{m1}^{*}\right\| \left\|\boldsymbol{X}_{m1}\t\bE_{m1}+\X_{m1}\t\X_{m1}(\bT_{0m1}-\bT_{m1}^t)\right\| \\
& \geq -\frac{1}{M} \sum_{m=1}^{M}\frac{1}{n_m}\left\|\bT_{m1}-\bT_{m1}^{*}\right\|
\left( \left\|\boldsymbol{X}_{m1}\t\bE_{m1}\right\|+\left\|\X_{m1}\t\X_{m1}(\bT_{0m1}-\bT_{m1}^t)\right\| \right) \\
& \geq -\frac{1}{M} \sum_{m=1}^{M}\frac{1}{n_m}\left\|\bT_{m1}-\bT_{m1}^{*}\right\|
\left( \sqrt{ 1/c_{1}} \sqrt{s_kq\ln (s_kqn)} +  \tau n_m\sqrt{s_kq/N_{\min }}\right) \\
& = - \sum_{m=1}^{M}\left\|\bT_{m1}-\bT_{m1}^{*}\right\|
\left( \sqrt{ 1/c_{1}} \frac{\sqrt{s_kq\ln (s_kqn)}}{Mn_m} +  \tau \frac{\sqrt{s_kq}}{M\sqrt{N_{\min }}}\right) \\
\end{align*}

Thus there is an event $E_{2}$ such that $\operatorname{Pr}\left(E_{2}^{C}\right) \leq 2M / n$. Conditions (C1) and (C3) (iv) mean that $\tau \sqrt{s_kq / N_{\min }} \leq \lambda_1\left|\mathcal{G}_{\min }\right|$. Hence in the event $E_{2}$,

\begin{align}
\Gamma_{1} & \geq -\sum_{j=1}^{p} \sum_{m=1}^{M} \left( \sqrt{ 1/c_{1}} \frac{\sqrt{s_kq\ln (s_kqn)}}{Mn_m} +  \tau \lambda_1\left|\mathcal{G}_{\min }\right|\right)\left\|\bT_{m(j)}-\bT_{m(j)}^{*}\right\|\\
& \geq -\sum_{j=1}^{p} \sum_{s=1}^{S_{j}} \sum_{\left\{m, m^{\prime} \in \mathcal{G}_{s}^{j}, m<m^{\prime}\right\}} \left( \sqrt{ 1/c_{1}} \frac{\sqrt{s_kq\ln (s_kqn)}}{M} +  \tau \lambda_1\left|\mathcal{G}_{\min }\right|\right) /\left|\mathcal{G}_{s}^{j}\right|\left\|\bT_{m(j)}-\bT_{m^{\prime}(j)}\right\|
\end{align}

Let $t_{n}=o(1)$, then for MCP penalty, $p_{\lambda_2}^{\prime}\left(4 t_{n}\right) \rightarrow \lambda_2$. Therefore, by Condition (C5) (i)
$$
\begin{array}{l}
\ell_{p}(\bT)-\ell_{p}\left(\bT^{*}\right) = \Gamma_{1}+\Gamma_{2}+\Gamma_{3} \\
\geq \sum_{j=1}^{p} \sum_{s=1}^{S_{j}} \sum_{\left\{m, m^{\prime} \in \mathcal{G}_{s}^{j}, m<m^{\prime}\right\}}\left\{\lambda_2-2 \lambda_1\sqrt{q} /\left|\mathcal{G}_{s}^{j}\right|-\left( \sqrt{ 1/c_{1}} \frac{\sqrt{s_kq\ln (s_kqn)}}{M} +  \tau \lambda_1\left|\mathcal{G}_{\min }\right|\right) /\left|\mathcal{G}_{s}^{j}\right|\right\}\left\|\bT_{m (j)}-\bT_{m^{\prime} (j)}\right\| \\
I\left(\bT_{0 m (j)} \neq 0 \text { or } \bT_{0 m^{\prime} (j)} \neq 0\right) \\
\geq \sum_{j=1}^{p} \sum_{s=1}^{S_{j}} \sum_{\left\{m, m^{\prime} \in \mathcal{G}_{s}^{j}, m<m^{\prime}\right\}}\left\{\lambda_2-\left(2 \lambda_1\sqrt{q}+ \sqrt{ 1/c_{1}} \frac{\sqrt{s_kq\ln (s_kqn)}}{M} +  \tau \lambda_1\left|\mathcal{G}_{\min }\right|\right) /\left|\mathcal{G}_{s}^{j}\right|\right\}\left\|\bT_{m (j)}-\bT_{m^{\prime} (j)}\right\| \\
I\left(\bT_{0 m (j)} \neq 0 \text { or } \bT_{0 m^{\prime} (j)} \neq 0\right) \\
\geq 0
\end{array}
$$
This completes the proof.

\section{Proof of Theorem 5}%\ref{Xnor}}
we prove the asymptotic normality of $\widetilde{\bX}$ 

$$
M^{-1} \mathbb{X}_{1}^{\top}\left(\Y-\mathbb{X}_{1} \widetilde{\bX}_{\mathcal{K} 1}\right)-\bbT_1(\widetilde{\bX}_{\mathcal{K} 1})\widetilde{\bX}_{\mathcal{K} 1}=0
$$

Hence,

$$
\widetilde{\bX}_{\mathcal{K} 1}-\bX_{0 \mathcal{K} 1}=\left(\mathbb{X}_{1}^{\top} \mathbb{X}_{1}\right)^{-1} \mathbb{X}_{1}^{\top} \bE-M\left(\mathbb{X}_{1}^{\top} \mathbb{X}_{1}\right)^{-1} \bbT_1(\widetilde{\bX}_{\mathcal{K} 1})\widetilde{\bX}_{\mathcal{K} 1}
$$

Since $\widetilde{\bX}_{\mathcal{K} 1} \in \mathcal{N}_{0}$, we have

$$
\left\|\bbT_1(\widetilde{\bX}_{\mathcal{K} 1})\widetilde{\bX}_{\mathcal{K} 1}\right\| \leq s_k^{1 / 2} |\mathcal{G}_{\text{max}}|p_{\gamma}^{\prime}\left(d_{n}\right)=o\left(\frac{\sqrt{s_k q N_{\min}}}{M}\right)
$$

Then under (C4) (i) and by $N_{\min} \gg s_{k}^{3 / 4} n^{1 / 2} $, we obtain

$$
\left\|M\left(\mathbb{X}_{1}^{\top} \mathbb{X}_{1}\right)^{-1} \bbT_1(\widetilde{\bX}_{\mathcal{K} 1})\widetilde{\bX}_{\mathcal{K} 1}\right\| \leq o\left(\frac{M}{N_{\min}}\frac{ \sqrt{s_k q N_{\min}}}{M}\right)=o\left(\sqrt\frac{s_k q}{N_{\min}}\right)
$$

Therefore,

$$
\widetilde{\bX}_{\mathcal{K} 1}-\widetilde{\bX}_{0 \mathcal{K} 1}=\left(\mathbb{X}_{1}^{\top} \mathbb{X}_{1}\right)^{-1} \mathbb{X}_{1}^{\top} \bE+o_{p}\left(\sqrt\frac{s_k q}{N_{\min}}\right)
$$

Thus, by Slutsky's lemma, to show the asymptotic distribution of $\widetilde{\bX}_{\mathcal{K} 1}-\widetilde{\bX}_{0 \mathcal{K}}$, it suffices to prove that of $\left(\mathbb{X}_{1}^{-1} \mathbb{X}_{1}\right)^{-1} \mathbb{X}_{1}^{\top} \bE$. For any given $a_{n}(s_k \times 1)$ and $c_{n}(q \times 1)$ , we consider the asymptotic distribution of the linear combination $a_{n}^{\top}\left(\widetilde{\bX}_{\mathcal{K} 1}-\widetilde{\bX}_{0 \mathcal{K} 1}\right) c_{n}$ i.e., $a_{n}^{\top}\left(\mathbb{X}_{1}^{\top} \mathbb{X}_{1}\right)^{-1} \mathbb{X}_{1}^{\top} \bE c_n.$
Since

$$
a_{n}^{\top}\left(\mathbb{X}_{1}^{\top} \mathbb{X}_{1}\right)^{-1} \mathbb{X}_{1}^{\top} \bE c_n = \sum_{i=1}^{n}a_{n}^{\top}\left(\mathbb{X}_{1}^{\top} \mathbb{X}_{1}\right)^{-1}  \mathbb{X}_{1(i)} (\bE_{(i)}\t c_n)=\sum_{i=1}^{n}a_{n}^{\top}\left(\mathbb{X}_{1}^{\top} \mathbb{X}_{1}\right)^{-1}  \mathbb{X}_{1(i)} \e^{\prime}_{i}
$$

where $\e^{\prime}_{i} = \bE_{(i)}\t c_n$ and $\mathbb{X}_{1(i)}$ is the $i$th row of $\mathbb{X}_{1}$. Note that $\mathrm{E}\e^{\prime}_{i} = 0 $, and $\operatorname{var}(\e^{\prime}_{i}) = \sigma^2$, so we have $\sigma_{n}^{2}(a_{n})=\operatorname{var}\left\{a_{n}^{\top}\left(\mathbb{X}_{1}^{\top} \mathbb{X}_{1}\right)^{-1} \mathbb{X}_{1}^{\top} \bE c_n\right\}=\sigma^{2} a_{n}^{\top}\left(\mathbb{X}_{1}^{\top} \mathbb{X}_{1}\right)^{-1} a_{n}>O(1 / n)$. Next we verify the Lindeberg-Feller condition. Under Conditions (C2) and (C4) (iii)

Since $E\left(\epsilon^{4}\right)<c$ for some constant $c>0$ by Condition 2, then

$$
\sum_{i=1}^{n}\left[\mathrm{E}\left(a_{n}^{\top}\left(\mathbb{X}_{1}^{\top} \mathbb{X}_{1}\right)^{-1}  \mathbb{X}_{1(i)} \e^{\prime}_{i}\right)^{4}\right]^{1 / 2} = \sum_{i=1}^{n}\left(a_{n}^{\top}\left(\mathbb{X}_{1}^{\top} \mathbb{X}_{1}\right)^{-1}  \mathbb{X}_{1(i)}\right)^{2} (\mathrm{E}\e^{\prime 4}_{i})^{1 / 2} 
$$

In fact, for any $i$, we have
\begin{align*}
\mathrm{E}\e^{\prime 4}_{i} =  \mathrm{E}(\e_{(i)}\t c_n)^4 = \mathrm{E}(\sum_{j=1}^{s_k}\e_{(i)j} c_{nj})^4 = C_4^2 \sum_{k \ne j}c_{nj}^2c_{nk}^2\mathrm{E}\left(\e_{(i)j}^2\e_{(i)k}^2\right) + \sum_{j}c_{nj}^4\mathrm{E}\left(\e_{(i)j}^4\right) \le 3 \sigma_{n}^4 s_{k}
\end{align*}

For some constant $c^{\prime}>0$, we have

$$
\begin{aligned}
& \max _{i} P\left(\left|a_{n}^{\top}\left(\mathbb{X}_{1}^{\top} \mathbb{X}_{1}\right)^{-1}  \mathbb{X}_{1(i)} \e^{\prime}_{i}\right|>\epsilon \sigma_{n}\right) 
\leq  \max _{i} \mathrm{E}\left(a_{n}^{\top}\left(\mathbb{X}_{1}^{\top} \mathbb{X}_{1}\right)^{-1}  \mathbb{X}_{1(i)} \e^{\prime}_{i}\right)^{2} /\left\{\epsilon^{2} \sigma_{n}^{2}\right\}\\
&\leq  \max _{i} \|a_{n}\|^2 \|\left(\mathbb{X}_{1}^{\top} \mathbb{X}_{1}\right)^{-1} \|^2 \|\mathbb{X}_{1(i)}\|^2 \mathrm{E}\e^{\prime 2}_{i} /\left\{\epsilon^{2} \sigma_{n}^{2}\right\}
\leq  \frac{s_k}{\epsilon^{2} \sigma_n^2 N_{\min}^2}
\end{aligned}
$$

% $$
% \begin{aligned}
% \mathrm{E}\left(\left\|\xi_{i}\right\|^{4}\right) & =\mathrm{E}\left\{\epsilon_{i}^{\top} \mathbb{X}_{1 i}^{\top}\left(\mathbb{X}_{1}^{\top} \mathbb{X}_{1}\right)^{-1} a_{n} a_{n}^{\top}\left(\mathbb{X}_{1}^{\top} \mathbb{X}_{1}\right)^{-1} \mathbb{X}_{1 i} \epsilon_{i}\right\}^{2} \leq\left\|a_{n}\right\|^{4}\left\|\left(\mathbb{X}_{1}^{\top} \mathbb{X}_{1}\right)^{-1}\right\|^{4} \mathrm{E}\left(\epsilon_{i}^{4}\right)\left\|\mathbb{X}_{1 i}\right\|^{4}=O\left(q^{2} / N_{\min }^{4}\right) \\
% \operatorname{Pr}\left(\left\|\xi_{i}\right\|>\varepsilon \sigma_{n}\left(a_{n}\right)\right) & \leq \mathrm{E}\left(\left\|\xi_{i}\right\|^{2}\right) /\left(\varepsilon^{2} \sigma_{n}^{2}\left(a_{n}\right)\right)=O\left(n q / N_{\min }^{2}\right)
% \end{aligned}
% $$

Therefore by condition (C5)(ii),
\begin{align*}
&\sigma_{n}^{-2} \sum_{i=1}^{n} \mathrm{E}\left[\left( a_{n}^{\top}\left(\mathbb{X}_{1}^{\top} \mathbb{X}_{1}\right)^{-1}  \mathbb{X}_{1(i)} \e^{\prime}_{i}\right)^{2} I\left(\left|a_{n}^{\top}\left(\mathbb{X}_{1}^{\top} \mathbb{X}_{1}\right)^{-1}  \mathbb{X}_{1(i)} \e^{\prime}_{i}\right|>\varepsilon \sigma_{n}\left(a_{n}\right)\right)\right]\\
& = \sigma_{n}^{-2} \sum_{i=1}^{n} \left\{\mathrm{E}\left( a_{n}^{\top}\left(\mathbb{X}_{1}^{\top} \mathbb{X}_{1}\right)^{-1}  \mathbb{X}_{1(i)} \e^{\prime}_{i}\right)^{4}\right\}^{1/2} \cdot \max_{i}\left\{\operatorname{Pr}\left(\left|a_{n}^{\top}\left(\mathbb{X}_{1}^{\top} \mathbb{X}_{1}\right)^{-1}  \mathbb{X}_{1(i)} \e^{\prime}_{i}\right|>\varepsilon \sigma_{n}\left(a_{n}\right)\right)^{1/2}\right\}\\
&\leq O\left(\sigma_{n}^{-2} \sigma_{n}^{2}s_k^{1/2} \frac{s_k}{\epsilon^{2} \sigma_n^2 N_{\min}^2}\right) \leq O\left(s_k^{3 / 2} n / N_{\min }^{2}\right) \rightarrow 0    
\end{align*}

This completes the proof.